%% file: main.tex
\documentclass[twoside,twocolumn,9pt]{article}
\usepackage{extsizes}
\usepackage[super,sort&compress,comma]{natbib} 
\usepackage[version=3]{mhchem}
\usepackage[left=1.5cm, right=1.5cm, top=1.785cm, bottom=2.0cm]{geometry}
\usepackage{balance}
\usepackage{mathptmx}
\usepackage{sectsty}
\usepackage{graphicx} 
\usepackage{adjustbox}
\usepackage{caption}
\usepackage{subcaption}
\usepackage{pifont}
\newcommand{\xmark}{\text{\ding{55}}}
\usepackage{lastpage}
\usepackage[format=plain,justification=justified,singlelinecheck=false,font={stretch=1.125,small,sf},labelfont=bf,labelsep=space]{caption}
\usepackage{float}
\usepackage{fancyhdr}
\usepackage{fnpos}
\usepackage[english]{babel}
\addto{\captionsenglish}{%
  \renewcommand{\refname}{Notes and references}
}
\usepackage{array}
\usepackage{droidsans}
\usepackage{charter}
\usepackage[T1]{fontenc}
\usepackage[usenames,dvipsnames]{xcolor}
\usepackage{setspace}
\usepackage{amssymb}
\usepackage[compact]{titlesec}
\usepackage{hyperref}
\usepackage{orcidlink}

\usepackage{tabularx,booktabs,array,multirow}
\usepackage{spverbatim}
\newcolumntype{Y}{>{\raggedright\arraybackslash}X}
\newcolumntype{C}[1]{>{\centering\arraybackslash}m{#1}} 
\usepackage{draftwatermark}

\usepackage{epstopdf}%This line makes .eps figures into .pdf - please comment out if not required.

\usepackage{credits}

\definecolor{cream}{RGB}{222,217,201}

\begin{document}

\pagestyle{fancy}
\thispagestyle{plain}
\fancypagestyle{plain}{
%%%HEADER%%%
\renewcommand{\headrulewidth}{0pt}
}
%%%END OF HEADER%%%

%%%PAGE SETUP - Please do not change any commands within this section%%%
\makeFNbottom
\makeatletter
\renewcommand\LARGE{\@setfontsize\LARGE{15pt}{17}}
\renewcommand\Large{\@setfontsize\Large{12pt}{14}}
\renewcommand\large{\@setfontsize\large{10pt}{12}}
\renewcommand\footnotesize{\@setfontsize\footnotesize{7pt}{10}}
\makeatother

\renewcommand{\thefootnote}{\fnsymbol{footnote}}
\renewcommand\footnoterule{\vspace*{1pt}% 
\color{cream}\hrule width 3.5in height 0.4pt \color{black}\vspace*{5pt}} 
\setcounter{secnumdepth}{5}

\makeatletter 
\renewcommand\@biblabel[1]{#1}            
\renewcommand\@makefntext[1]% 
{\noindent\makebox[0pt][r]{\@thefnmark\,}#1}
\makeatother 
\renewcommand{\figurename}{\small{Fig.}~}
\sectionfont{\sffamily\Large}
\subsectionfont{\normalsize}
\subsubsectionfont{\bf}
\setstretch{1.125} %In particular, please do not alter this line.
\setlength{\skip\footins}{0.8cm}
\setlength{\footnotesep}{0.25cm}
\setlength{\jot}{10pt}
\titlespacing*{\section}{0pt}{4pt}{4pt}
\titlespacing*{\subsection}{0pt}{15pt}{1pt}
%%%END OF PAGE SETUP%%%

%%%FOOTER%%%
\fancyfoot{}
\fancyfoot[LO,RE]{\vspace{-7.1pt}\includegraphics[height=9pt]{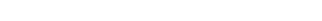}}
\fancyfoot[CO]{\vspace{-7.1pt}\hspace{13.2cm}\includegraphics{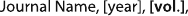}}
\fancyfoot[CE]{\vspace{-7.2pt}\hspace{-14.2cm}\includegraphics{head_foot/RF}}
\fancyfoot[RO]{\footnotesize{\sffamily{1--\pageref{LastPage} ~\textbar  \hspace{2pt}\thepage}}}
\fancyfoot[LE]{\footnotesize{\sffamily{\thepage~\textbar\hspace{3.45cm} 1--\pageref{LastPage}}}}
\fancyhead{}
\renewcommand{\headrulewidth}{0pt} 
\renewcommand{\footrulewidth}{0pt}
\setlength{\arrayrulewidth}{1pt}
\setlength{\columnsep}{6.5mm}
\setlength\bibsep{1pt}
%%%END OF FOOTER%%%

%%%FIGURE SETUP - please do not change any commands within this section%%%
\makeatletter 
\newlength{\figrulesep} 
\setlength{\figrulesep}{0.5\textfloatsep} 

\newcommand{\topfigrule}{\vspace*{-1pt}% 
\noindent{\color{cream}\rule[-\figrulesep]{\columnwidth}{1.5pt}} }

\newcommand{\botfigrule}{\vspace*{-2pt}% 
\noindent{\color{cream}\rule[\figrulesep]{\columnwidth}{1.5pt}} }

\newcommand{\dblfigrule}{\vspace*{-1pt}% 
\noindent{\color{cream}\rule[-\figrulesep]{\textwidth}{1.5pt}} }

\makeatother
%%%END OF FIGURE SETUP%%%

%%%TITLE, AUTHORS AND ABSTRACT%%%
\twocolumn[
  \begin{@twocolumnfalse}
{\includegraphics[height=30pt]{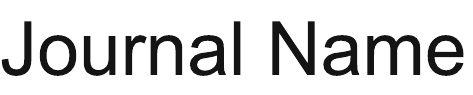}\hfill\raisebox{0pt}[0pt][0pt]{\includegraphics[height=55pt]{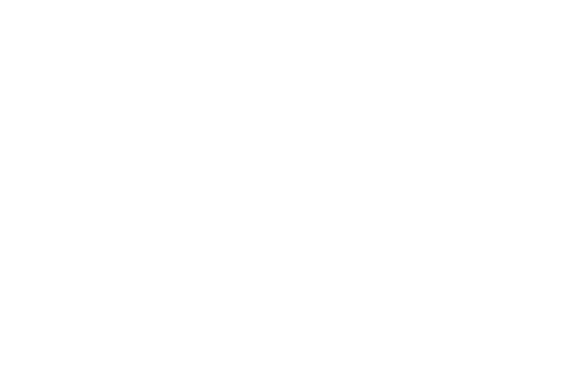}}\\[1ex]
\includegraphics[width=18.5cm]{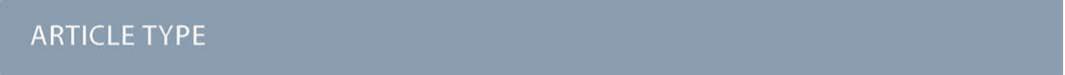}}\par
\vspace{1em}
\sffamily
\begin{tabular}{m{4.5cm} p{13.5cm} }

\includegraphics{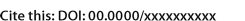} & \noindent\LARGE{\textbf{PRISM: Protocol Refinement through Intelligent Simulation Modeling$^\dag$}} \\
\vspace{0.3cm} & \vspace{0.3cm} \\

& \noindent\large
{Brian Hsu~\orcidlink{0009-0004-5027-607X}~$^{\dag,a,b}$,
 Priyanka V Setty~\orcidlink{0000-0002-3373-5981}~$^{\dag,a,b}$,
 Rory M Butler~\orcidlink{0000-0003-1667-1709}~$^{\dag,a,b}$,
 Ryan Lewis~\orcidlink{0000-0002-3000-2811}~$^{a}$,
 Casey Stone~\orcidlink{0009-0006-9208-4046}~$^{a}$,
 Rebecca Weinberg~\orcidlink{0000-0003-1398-0039}~$^{a}$,
 Thomas Brettin~\orcidlink{0000-0001-9301-9760}~$^{a,b}$,
 Rick Stevens~\orcidlink{0000-0002-4268-4020}~$^{a,b}$,
 Ian Foster~\orcidlink{0000-0003-2129-5269}~$^{a,b}$,
 Arvind Ramanathan~\orcidlink{0000-0002-1622-5488}~$^{a,\ast}$
}\\

\includegraphics{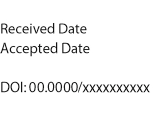} & \noindent\normalsize{\input{sections/abstract}} \\

\end{tabular}

 \end{@twocolumnfalse} \vspace{0.6cm}

  ]
%%%END OF TITLE, AUTHORS AND ABSTRACT%%%

%%%FONT SETUP - please do not change any commands within this section
\renewcommand*\rmdefault{bch}\normalfont\upshape
\rmfamily
\section*{}
\vspace{-1cm}

\newcommand{\ian}[1]{{\textcolor{red}{Ian: #1 }}}

%%%FOOTNOTES%%%

% \footnotetext{\textit{$^{a}$~Address, Address, Town, Country. Fax: XX XXXX XXXX; Tel: XX XXXX XXXX; E-mail: ramanathana@anl.gov}}
% \footnotetext{\textit{$^{b}$~Address, Address, Town, Country. }}

% Affiliation A (Argonne)
\footnotetext{\textit{$^{a}$~Argonne National Laboratory, 9700 S. Cass Avenue, Lemont, IL 60439, USA.}}

% Affiliation B (UChicago)
\footnotetext{\textit{$^{b}$~Department of Computer Science, University of Chicago, Chicago, IL 60637, USA.}}

% Corresponding Author Email
\footnotetext{\textit{$^{\ast}$~E-mail: ramanathana@anl.gov}}

%Please use \dag to cite the ESI in the main text of the article.
%If you article does not have ESI please remove the the \dag symbol from the title and the footnotetext below.
% \footnotetext{\textcolor{red}{\dag~Supplementary Information available: [details of any supplementary information available should be included here]. See DOI: 00.0000/00000000.}}
%additional addresses can be cited as above using the lower-case letters, c, d, e... If all authors are from the same address, no letter is required

\footnotetext{\dag~These authors contributed equally.}
%Additional footnotes to the title and authors can be included \textit{e.g.}\ `Present address:' or `These authors contributed equally to this work' as above using the symbols: \ddag, \textsection, and \P. Please place the appropriate symbol next to the author's name and include a \texttt{\textbackslash footnotetext} entry in the the correct place in the list.}

%%%END OF FOOTNOTES%%%

%%%MAIN TEXT%%%%

\input{sections/introduction}
\input{sections/methods}
\input{sections/results}
\input{sections/conclusions}

\section*{Author contributions}
%\textcolor{red}{We strongly encourage authors to include author contributions and recommend using \href{https://casrai.org/credit/}{CRediT} for standardised contribution descriptions. Please refer to our general \href{https://www.rsc.org/journals-books-databases/journal-authors-reviewers/author-responsibilities/}{author guidelines} for more information about authorship.}

% The CRediT taxonomy
% - Conceptualization
% - Data curation
% - Formal analysis
% - Funding acquisition
% - Investigation
% - Methodology
% - Project administration
% - Resources
% - Software
% - Supervision
% - Validation
% - Visualization
% - Writing -- original draft
% - Writing -- review \& editing

\credit{B.H.}{Investigation,Methodology,Software}
\credit{P.S.}{Investigation,Methodology,Writing -- original draft,Writing -- review \& editing,Software}
\credit{R.B.}{Investigation,Methodology,Writing -- original draft,Writing -- review \& editing,Software}
\credit{R.L.}{Investigation}
\credit{C.S.}{Investigation}
\credit{R.W.}{Investigation}
\credit{T.B.}{Investigation}
\credit{R.S.}{Conceptualization,Resources,Funding acquisition}
\credit{I.F.}{Writing -- review \& editing,Funding acquisition}
\credit{A.R.}{Conceptualization,Resources,Writing -- review \& editing,Funding acquisition}

\insertcreditsstatement

\section*{Conflicts of interest}
There are no conflicts of interest to declare.

\section*{Data availability}
% \textcolor{red}{A data availability statement (DAS) is required to be submitted alongside all articles. Please read our} \href{https://www.rsc.org/journals-books-databases/author-and-reviewer-hub/authors-information/prepare-and-format/data-sharing/#dataavailabilitystatements}{full guidance on data availability statements} \textcolor{red}{for more details and examples of suitable statements you can use.}
The code and prompts for PRISM can be found at \href{https://github.com/ramanathanlab/PRISM}{https://github.com/ramanathanlab/PRISM}.

\section*{Acknowledgements}
% \textcolor{red}{Contributors (those not already included as co-authors) may be acknowledged. Keep this section as brief as possible. You should also declare all sources of funding and any use of artificial intelligence (AI) tools at this point.}
We acknowledge support from the US Department of Energy's Biological and Environmental Research (BER) and Advanced Scientific Computing Research (ASCR), DE-FOA-0003600. Various versions of Anthropic's Claude, Google's Gemini, and OpenAI's GPT models were used to assist in writing this manuscript's text.

%%%END OF MAIN TEXT%%%


\balance

\renewcommand\refname{References}

%%%REFERENCES%%%
\bibliography{rsc} %You need to replace "rsc" on this line with the name of your .bib file
\bibliographystyle{rsc} %the RSC's .bst file

\clearpage
\input{sections/appendix}

\end{document}

%% file: sections/abstract.tex
Automating experimental protocol design and execution remains as a fundamental bottleneck in realizing self-driving laboratories. We introduce PRISM (Protocol Refinement through Intelligent Simulation Modeling), a framework that automates the design, validation, and execution of experimental protocols on a laboratory platform composed of off-the-shelf robotic instruments. PRISM uses a set of language-model-based agents that work together to generate and refine experimental steps. The process begins with automatically gathering relevant procedures from web-based sources describing experimental workflows. These are converted into structured experimental steps (e.g., liquid handling steps, deck layout and other related operations) through a planning, critique, and validation loop. The finalized steps are translated into the Argonne MADSci protocol format, which provides a unified interface for coordinating multiple robotic instruments (Opentrons OT-2 liquid handler, PF400 arm, Azenta plate sealer and peeler) without requiring human intervention between steps. To evaluate protocol-generation performance, we benchmarked both single reasoning models and multi-agent workflow across constrained and open-ended prompting paradigms. The resulting protocols were validated in a digital-twin environment built in NVIDIA Omniverse to detect physical or sequencing errors before execution. Using Luna qPCR amplification and Cell Painting as case studies, we demonstrate PRISM as a practical end-to-end workflow that bridges language-based protocol generation, simulation-based validation, and automated robotic execution.

%% file: sections/introduction.tex
\section{Introduction}

%\subsection{Problem Statement and Motivation}

The automation of experimental protocol design and execution remains a fundamental bottleneck in realizing self-driving laboratories. While robotic platforms are increasingly available, translating scientific intent into executable laboratory protocols requires substantial domain expertise, as the process is error-prone and demands intimate knowledge of both experimental procedures and instrument-specific formatting requirements. Existing protocol description languages provide structured representations but remain static and hardware-specific, requiring manual rewriting when laboratory configurations change. Recent work has shown that large language models (LLMs) can generate plausible experimental procedures from natural language descriptions, yet these outputs often contain parameter underspecification, physical infeasibility, or sequencing errors that would cause failures during execution. Meanwhile, simulation and digital twin technologies have advanced laboratory monitoring and documentation but have not been integrated into the protocol generation pipeline as effective pre-execution validation tools. Direct testing of unvalidated protocols on physical hardware is costly, time-consuming, and risks both equipment damage and material waste. These challenges motivate the development of systems that combine automated protocol generation with rigorous pre-execution validation to enable safe, reliable, and fully autonomous laboratory operation.

\subsection{Protocol Languages and Lab Automation Standards}

Early efforts to formalize laboratory procedures focused on creating machine-readable descriptions of experimental steps. XDL \cite{Steiner2019} demonstrated this for automated chemistry by expressing operations such as heating or adding reagents in a hardware-agnostic language. Autoprotocol \cite{Autoprotocol2025} extended this idea to cloud-operated labs more broadly, introducing a JSON-based specification for generic laboratory actions like pipetting, incubating, sealing, and measuring so that protocols could be executed across diverse automation systems. More integrated platforms such as Aquarium \cite{Vrana2021Aquarium} coupled protocol specification with inventory tracking and execution oversight, using a high-level language and Laboratory Information Management System (LIMS) to support technician-in-the-loop workflows.

While these systems illustrate a clear progression toward more structured and reproducible experimentation, they are all limited in important respects. In particular, protocol descriptions are static, tied to specific configurations, and offer little support for validating physical feasibility before execution. They also require specialized technical expertise to develop and adapt. These gaps motivate the need for more adaptive approaches that combine structured representations with contextual reasoning and automated validation, which we explore in this work.

\subsection{Direct LLM-Guided Protocol Generation}

% \ian{It seems that we do not cite \cite{boiko2023autonomous}, although the tech report version is in the bib file. That needs to be included and discussed.}

% \ian{Some other papers that we may want to cite: LLMs in chemistry \cite{ramos2025review}, survey of LLM-based scientific agents (perhaps too far afield?) \cite{ren2025towards}, automation and LLMs in catalysis \cite{su2024automation}, LLMs generate robotic scripts \cite{inagaki2023llms}, LLMs and robotics \cite{karli2024alchemist}, LLMs and lab safety \cite{zhou2024labsafety}, LLMs and chemical synthesis \cite{ruan2024automatic}, LLMs in science \cite{zhang2024comprehensive}, AI-powered robotic chemist \cite{song2025multiagent}, }

LLMs have shown the ability to generate multi-step experimental workflows when used alongside external tools. Recent surveys have cataloged over 260 such ``scientific LLMs'' across disciplines \cite{zhang2024comprehensive}, classifying them as ``autonomous agents'' that combine reasoning, planning, and tool use to solve complex scientific tasks \cite{ramos2025review, ren2025towards}. ChemCrow \cite{ChemCrow} demonstrated LLM integration in chemistry workflows by coupling an LLM with reaction databases and planners, allowing high-level goals to be decomposed into executable steps. Coscientist \cite{boiko2023autonomous} advanced further by integrating documentation search to autonomously execute reactions. 

More recently, hierarchical multi-agent systems like ChemAgents \cite{song2025multiagent} and LLM-RDF \cite{ruan2024automatic} have demonstrated end-to-end chemical discovery, though they notably rely on human-in-the-loop mechanisms or manual verification to ensure safety prior to execution. Building on this idea, BioPlanner \cite{o2023bioplanner} examined whether similar capabilities extend to biological protocols, evaluating the completeness and clarity of model-generated procedures and conducting limited feasibility tests. Similarly, Inagaki et al. \cite{inagaki2023llms} and Alchemist \cite{karli2024alchemist} utilized LLMs to generate executable robotic scripts (e.g., for Opentrons), demonstrating that while syntax errors can be caught via standard API simulators, physical execution failures persist without robust environmental simulation. 

Yet more recent work, such as ProtoCode \cite{jiang2024protocode}, moves toward making outputs actionable by introducing structured intermediate representations that bridge free-text descriptions and machine-interpretable robotic actions. Agentic Lab \cite{wang2025agentic} employs multi-agent orchestration for knowledge retrieval, protocol composition, and iterative refinement through LLM-based self-reflection and human feedback. 

Together, these efforts reflect a growing interest in using language models not only to outline experimental intent but also to guide downstream automation systems.
Yet despite this progress, limitations remain. First, parameter underspecification is common: model outputs often omit key details such as reagent volumes, concentrations, incubation times, or mixing cycles, which are essential for execution. Second, some instructions are physically infeasible, for instance, suggesting movements of labware that exceed the reach of a robotic arm, stacking plates in ways incompatible with the deck layout, or repositioning items that cannot be grasped by the available hardware. In fact, recent benchmarking on laboratory safety has revealed that even state-of-the-art models fail to identify critical hazards in realistic scenarios over 30\% of the time, underscoring the risks of unverified execution \cite{zhou2024labsafety}. Third, there is a gap between text generation and robotic execution: most outputs remain in natural language or semi-structured pseudocode, requiring manual translation into the specific formats used by laboratory robots (e.g., YAML or JSON). Finally, current evaluations provide limited validation or feedback, relying mainly on expert review rather than simulation or digital-twin testing. BioPlanner includes small-scale experimental verification, but such validation remains the exception. Agentic Lab provides real-time feedback during human physical execution through AR glasses and vision-language models. However, this approach validates procedures during execution rather than simulating potential outcomes beforehand.

Together, these findings suggest that while LLMs can assist in translating high-level scientific intent into procedural steps, their outputs still need systematic refinement and physical validation. Bridging this gap requires combining language generation with simulation, contextual reasoning, and platform-aware validation—an approach that motivates the PRISM framework developed in this work.

\begin{figure*}[ht]
    \centering
    \includegraphics[width=\textwidth]{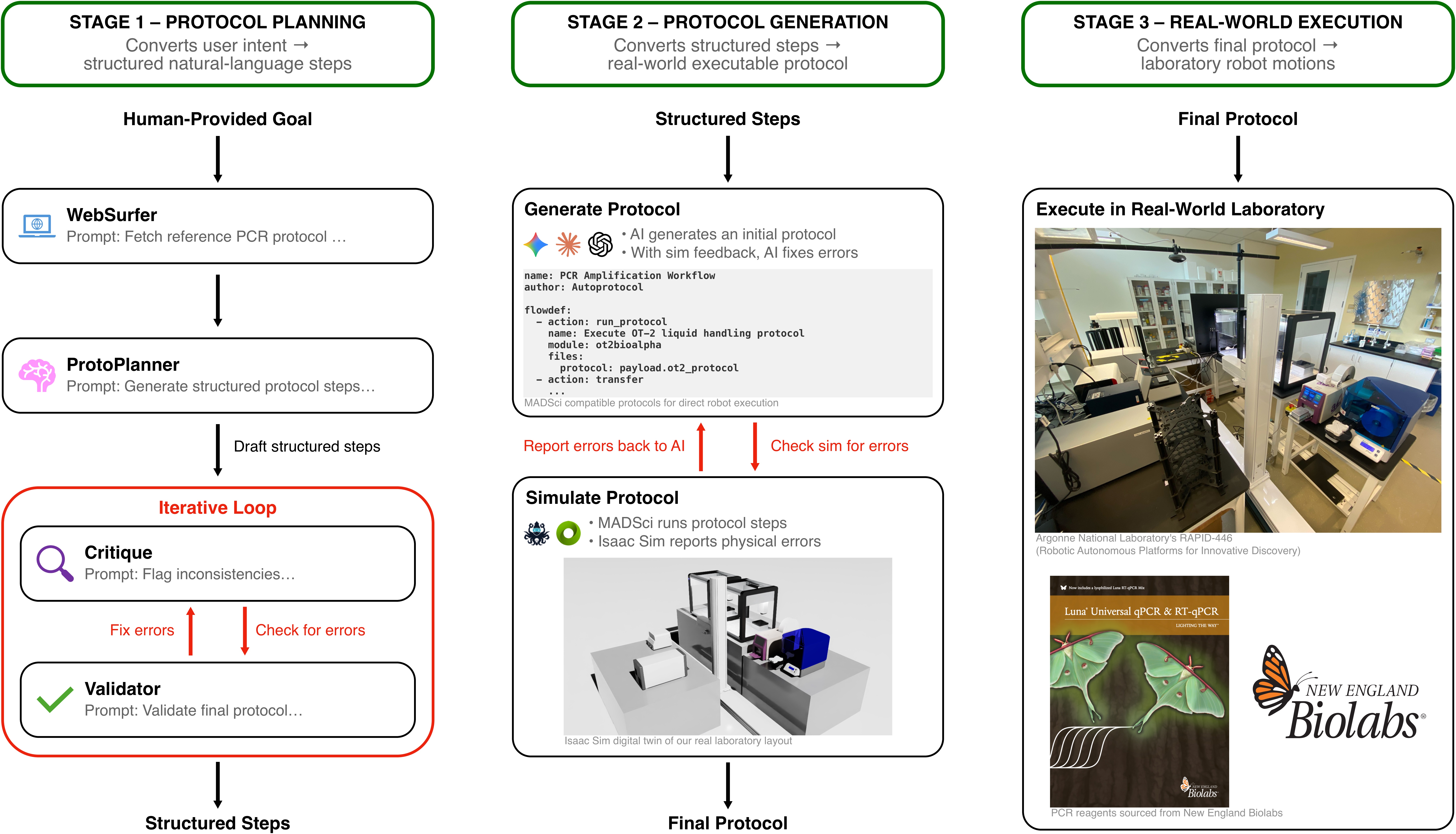}
    \caption{Overview of the PRISM framework for protocol generation and execution.
The system consists of three main stages: Protocol Planning, where user intent is converted into structured steps; Protocol Generation, where structured English instructions are transformed into robot-aware actions and iteratively refined through validation cycles in Omniverse before execution; and Real-World Execution, where the full pipeline is validated using the Luna qPCR protocol in our autonomous laboratory.}
    \label{fig:prism_overview}
\end{figure*}

\subsection{Simulations and Digital Twins}

Digital twins (in this context: virtual replicas of physical laboratory systems) provide opportunities to design, test, deploy, monitor, and control real-world robotic processes \cite{douthwaite2021modular}. Such simulation environments enable offline testing and validation of robotic operations through physics-based collision detection, joint constraints, and realistic motion planning. However, most applications of digital twins for laboratory automation only emphasize post-hoc documentation, execution tracing, and real-time monitoring, rather than pre-execution validation and refinement. Recent reviews in catalysis have highlighted this gap, noting that while LLMs offer flexibility, the development of digital twins to ensure the physical safety and integrity of generated protocols remains in its infancy \cite{su2024automation}.

Previous advances in LLM applications have demonstrated certain key but limited pre-execution validation and refinement techniques. For example, CLAIRIFY \cite{yoshikawa2023large} developed an approach using verifier-assisted iterative prompting to generate valid programs in domain-specific languages. By combining automatic iterative prompting with syntax and rule-based program verification, CLAIRIFY ensures valid programs that incorporate environmental constraints. However, while this verification checks syntactic correctness and certain manually defined rules, it does not comprehensively validate physical feasibility, collision risks, or hardware constraints through high-fidelity simulation. ORGANA \cite{darvish2025organa}, building on CLAIRIFY, executes protocols directly on physical hardware, solving task and motion planning problems on the fly as part of the protocol's execution process. Similarly, Agentic Lab's real-time monitoring through vision-language models enables continuous learning and protocol refinement based on post-execution analysis, but it does not provide the grounded pre-execution validation that simulation-based approaches offer. These system's abilities to detect and correct errors depends on syntax verification, handcrafted rule-based verification, and real-time visual observation during physical execution, rather than anticipating and preventing errors through prior simulation. Current approaches therefore do not leverage simulation's potential for pre-execution validation, executing AI-generated protocols directly on hardware despite the risks of equipment damage, material waste, and time costs.

\subsection{Contributions and Goals}

A critical gap exists in the current landscape of laboratory automation: no existing work combines end-to-end LLM-driven protocol generation and simulation-based error detection into a unified pipeline for fully autonomous generation of trustworthy protocols. Digital twin work provides comprehensive monitoring and tracing capabilities but remains insufficiently integrated with automated protocol generation systems.

The PRISM framework addresses this gap through a three-stage pipeline that integrates multi-agent LLM planning, reasoning-model-driven protocol generation, and physics-based simulation error detection. PRISM positions simulation as a mandatory pre-execution gate, creating a closed feedback loop where detected errors inform iterative protocol refinement. We demonstrate the system's effectiveness through PCR amplification (experimentally validated on our robotic platform) and Cell Painting (computationally validated), while providing comparative analysis across multiple state-of-the-art language models. This work represents a step toward fully autonomous experimental design and execution, bridging the divide between AI-generated plans and safe, reliable laboratory automation.

%% file: sections/methods.tex
\section{Methods}
We introduce PRISM: Protocol Refinement through Intelligent Simulation Modeling, a framework that automates the design, validation, and execution of experimental protocols on a laboratory platform composed of off-the-shelf robotic instruments. We use a multi-agent framework (Figure~\ref{fig:prism_overview}, Stage 1), described in Section~\ref{sec:Protocol Planning Methods}, to generate the steps of an experimental procedure of interest, initially described in plain English. The steps follow a structured format that the protocol generator recognizes and can convert into robot-compatible instructions. The protocol generator (Figure~\ref{fig:prism_overview}, Stage 2) described in Section~\ref{sec:Initial Protocol Generation} combines these structured steps with feedback from a digital twin simulation described in Section~\ref{sec:Iterative Error Detection in Simulation}, iteratively refining the design until a valid experimental protocol is produced. The final valid protocol is then ready to execute in the real-world laboratory (Figure~\ref{fig:prism_overview}, Stage 3) as described in Section~\ref{sec:Real World Execution}.

\subsection{Protocol Planning}\label{sec:Protocol Planning Methods}
The protocol planning stage focuses on generating complete structured liquid handling instructions in plain English that capture all the reagents, labware, and procedural details required to execute an experiment on an automated platform. These steps are not directly robot-executable, but are formatted to preserve the logical sequence of actions and physical constraints necessary for robotic translation.

We evaluate two complementary prompting paradigms, \textbf{constrained} and \textbf{open-ended}, for which the full prompt templates are provided in \autoref{sec:Appendix Protocol Planning}. In the constrained setting, models receive explicit reagent volumes, well layouts, and step-format rules, whereas open-ended prompting provides only high-level experimental goals, requiring models to infer volumes, mappings, and transfer sequences. We test each prompting paradigm across two architectural frameworks—a \textbf{multi-agent system} and a \textbf{single-agent system} to assess each model's ability to generate, refine, and self-correct experimental protocols over three iterative refinement cycles.

% \subsubsection{Prompt Paradigms}

% \textbf{Constrained prompting:} Provides explicit structural and contextual guidance, including deck layout, reagent constraints, and volume schemas. It enforces strict adherence to the OT-2’s physical rules (e.g., pipette limits, well spacing, master mix calculations).
% \textbf{Open-ended prompting:} Allows the model greater autonomy to reason through the protocol design, focusing on logical and biologically coherent generation while maintaining feasibility within robotic constraints.

\subsubsection{Multi- vs.\ Single-agent Protocol Planning Frameworks}

We compare two architectural approaches for protocol planning: multi-agent and single-agent. These two frameworks differ primarily in how reasoning is distributed and how errors are detected and corrected.

In the \textbf{multi-agent system}, the task is decomposed across four specialized agents, each with a distinct role. The \emph{WebSurfer agent} retrieves experimental information from online sources and extracts procedural steps and reagents. The \emph{Protocol Planner agent} then converts this unstructured description into a structured plan, selecting labware, assigning reagents to specific deck positions, and rewriting each action into a consistent liquid-handling schema. Next, a \emph{Critique agent} evaluates the structured steps for missing information, volume inconsistencies, formatting issues, and logical breakdowns. Finally, the \emph{Validator agent} applies any needed corrections and re-submits updated steps to the Critique agent, establishing an iterative refinement loop. Because each agent reasons only within its specialized scope, the system benefits from modularity: reasoning is constrained, errors are easier to isolate, and corrective feedback is more targeted. However, this setup introduces dependencies between agents, and failures in early stages can propagate unless caught by downstream checks.

In contrast, the \textbf{single-agent} (reasoning model) framework relies on a single model to perform all stages of the task within one continuous chain of thought. The model must retrieve relevant experimental details, infer appropriate deck layouts, assign reagents, compute volumes, and generate a fully enumerated set of structured liquid-handling steps in a single pass, without external critique or role decomposition. This approach benefits from simplicity and faster execution but places a significantly higher cognitive load on the model: all planning, validation, error detection, and correction must occur internally. Without explicit modular boundaries, the model may overlook inconsistencies, lose track of earlier constraints, or produce logically correct but physically invalid steps, especially as workflows become longer and more complex.

Overall, the multi-agent system distributes reasoning across specialized components that reinforce each other, whereas the single-agent system must manage the entire reasoning space at once. As we explain in more detail below, we find that the single-agent system works reasonably well for short protocols with few dependencies, such as PCR, where the sequence is largely fixed (prepare reactions → thermocycle) and the exact order of reagent additions is not critical. In contrast, workflows like Cell Painting have many interdependent steps where timing, order, and reagent interactions matter, making them more prone to cascading reasoning errors when handled by a single model. For these complex, more structured protocols, the multi-agent framework is more robust and consistently produces correct outputs.

\subsubsection{Evaluation Setup} 

We benchmark the two frameworks under constrained and open-ended prompting using five state-of-the-art LLMs: \textbf{GPT-5}\cite{gpt5}, \textbf{Claude Opus 4.1}\cite{claude_opus}, \textbf{Claude Sonnet 4.5} \cite{claude_sonnet}, \textbf{Gemini 2.5 Pro} \cite{gemini}, and \textbf{Gemini 2.5 Flash} \cite{gemini}.  We evaluated each model–framework combination under both constrained and open-ended prompting configurations for up to three refinement iterations. In cases where convergence was not achieved within three iterations, the final attempt was recorded for analysis. For the PCR experiment, the GPT-5 constrained multi-agent protocol was used as ground truth, as it produced a fully correct protocol without requiring any correction cycles and was validated by an experimental biologist. Each generated protocol was compared against this reference across predefined ground truth categories like logical transfers, master mix calculations, reagent and well assignments, pipette selection, and step formatting to identify correct, missing, and extraneous steps (true positives, false negatives, and false positives). F1 scores were then computed within and across these categories to quantify generation accuracy.

\subsection{Initial Protocol Generation}\label{sec:Initial Protocol Generation}

To translate from English protocols into YAML format, we used LLMs with reasoning capabilities that perform a thinking step before generating responses. We tested this stage of the framework with six reasoning models: \textbf{GPT-5} \cite{gpt5}, \textbf{Claude Opus 4.1} \cite{claude_opus}, \textbf{Claude Sonnet 4.5} \cite{claude_sonnet}, \textbf{Gemini 2.5 Pro} \cite{gemini}, \textbf{Gemini 2.5 Flash} \cite{gemini}, and \textbf{Gemini 2.5 Flash-Lite} \cite{gemini_flash_lite}.

All model generation parameters were kept at their default values for consistency and simplicity. When provided with input from the protocol planning stage, these models convert high-level English instructions into structured YAML output by breaking down the instructions into specific robot commands that can be executed by laboratory automation systems. The resulting YAML files contain sequentially ordered robot actions along with their associated parameters, which together form a complete machine-readable protocol.

The complexity of laboratory workflows makes simple rule-based translations from English to robot commands insufficient for many scenarios. What appears as a straightforward instruction in English, such as ``seal the plate,'' often requires multiple coordinated robot actions, including retrieving the plate from its current location, transferring it to a sealing robot, and executing the sealing operation. This complexity increases when working with robots that have specific operational requirements, such as those that must be opened before plates can be inserted and closed before operations can be performed, or when unique position and rotation requirements of the source and destination robots conflict. When laboratories incorporate multiple robots with different operational constraints and spatial restrictions, the sequence of actions becomes non-trivial to determine, particularly when synchronization between robots is necessary. A single misplaced step in the sequence can lead to physical impossibilities, such as attempting to place a plate in a closed device or trying to operate a robot before proper preparation. These challenges make reasoning about the entire workflow important, as the model must understand not only each individual robot's capabilities but also how they interact across the laboratory's spatial and operational constraints.

\subsubsection{Model Input}

The translation process uses three main input components to generate robot-executable protocols. First, the system receives English process steps with scientific parameters such as volumes, temperatures, and timing information derived from the previous protocol planning stage. These steps describe the high-level scientific procedures that need to be performed, such as the PCR and Cell Painting protocols examined in this work. Second, the system is provided with detailed information about the available robots' capabilities, including their possible actions, required arguments for each action, and unique operational constraints. Third, the system receives specifications for the expected output format, detailing the structure of the YAML file that will be used by the laboratory automation system (Argonne's MADSci \cite{madsci_JOSS} framework, building on earlier work~\cite{vescovi2023towards}) to execute the protocol. These documents can be used as separate files when working with tools like Claude Code, which allows for better composition and organization, or they can be concatenated into a single prompt when using models via direct API calls to the models as we did in this work.

The YAML output format follows the MADSci framework’s protocol specification, which encodes a protocol as an ordered sequence of atomic robot actions. Each step in the YAML file includes the  robot to use, the action to perform, and any arguments required for that action. This structured format breaks down high-level instructions like ``run PCR'' into a precise sequence of robot movements, plate transfers, and device operations. The YAML structure follows a consistent pattern that allows the laboratory automation system to interpret and execute each step sequentially, with appropriate error handling and reporting opportunities between steps.

Validation of the generated YAML files is handled by the MADSci framework, which performs syntax checking and basic capability verification. MADSci examines the YAML file to ensure that all required fields are present, that the robots referenced in the protocol exist in the laboratory configuration, and that the requested actions are within each robot's capabilities. This built-in validation eliminates the need for separate validation code in our system. MADSci also checks whether the arguments provided for each action match the expected inputs for the corresponding robot. The additional, more advanced validation that we introduce, such as collision detection and object presence verification, is handled during the simulation phase, where physical execution of the protocol is tested in a virtual environment.

Robot information is organized to make laboratory setup more flexible and reusable. Each robot's capabilities, actions, arguments, and unique operational requirements are defined once and packaged together. This approach allows researchers to define laboratory-specific restrictions separately from the intrinsic properties of the robots. When setting up a new laboratory configuration, researchers can simply specify which robots are being used and derive prompts from these pre-written descriptions, then add any laboratory-specific constraints without needing to redefine the robots' capabilities. This separation makes it easier to compose new laboratory layouts by reusing existing robot definitions and only adding the minimal information needed for the specific laboratory setup being modeled.

\subsection{Iterative Error Detection in Simulation}\label{sec:Iterative Error Detection in Simulation}

To build the digital twin environment for our simulation we acquired Computer Aided Design (CAD) models of the robots from vendors or public repositories. These models were initially static, lacking physics and joint information needed for movement simulation. We added necessary physics properties to each model by defining collision geometries, and placing joint connections at locations that match the real robots, a process which required measuring the physical robots' joint positions and mapping them to corresponding locations in the CAD models. We also developed simulation drivers that translate commands intended for the real robots into the appropriate control signals for moving the simulated joints in Omniverse \cite{omniverse}, allowing the digital robots to mimic the movements of their physical counterparts. These collision and joint additions enable the simulation to be a more useful model of the real world for the downstream error detection steps without requiring substantial extra effort by the researcher.

Setting up the laboratory scene involved measuring the real-world laboratory space and placing the digital robots in matching positions within the simulation. A CAD model's high fidelity allows for beyond millimeter-level accuracy in robot size and placement for future research in more flexible robot control. However, this implementation focused on functional accuracy rather than precise positioning as robot motions occurred only in predefined configurations. By placing the robots in a visually matching configuration, we enabled reliable collision detection during key events such as placing and retrieving plates from robots and simplified manual visual confirmation that protocol simulation was proceeding as expected.

\subsubsection{Sim-to-real Correspondence}

The communication between MADSci and the simulation environment uses custom robot interface modules that replace MADSci's default connections to physical robots with customized network messages to the digital twin over ZeroMQ. This approach allows MADSci to operate as if it were communicating with real robots while in actuality sending commands to the simulation. The Omniverse scripts receive these messages and direct the simulated robots to perform the requested actions, then send completion or error messages back to MADSci. This communication method avoids the complexities of replicating each robot's unique network protocol, leveraging the proven MADSci capabilities.

An important aspect of our approach is that MADSci behaves identically whether controlling real or simulated robots. We run the same MADSci software used in physical laboratories, with only the robot interface modules changed to communicate with the simulation. This means that all command validation, orchestration logic, and error handling built into MADSci operates exactly as it would in a real-world setting. The timing and synchronization of robot actions occur in real-time, with messages sent to and from the simulation as they would be with physical robots. This consistency gives confidence that protocols working correctly in simulation will behave similarly when transferred to the real laboratory, as the only difference is the replacement of physical robots with their simulated counterparts.

\subsubsection{Error Detection Feedback}

The simulation detects several types of errors through physics-based collision detection. We configure the system to ignore expected collisions, such as when a robot gripper intentionally contacts a plate that it is meant to pick up, while monitoring solely for unexpected collisions between robots or between materials and the environment. When an unintended collision occurs, the simulation reports it to MADSci with an English-language description of the problem (identifying which objects collided by name). This error reporting causes the MADSci orchestrator to terminate the experiment execution, just as it would if a similar error occurred in the physical laboratory. The complete execution log up to the point of failure can be captured, providing a record of all steps successfully completed before the error plus context about what was happening when the collision was detected.

Beyond collision detection, the simulation checks for object presence and validates command execution. A ray-casting system verifies whether plates or other materials are present when robots attempt to interact with them. If a robot tries to pick up a plate that is not at the expected location, the system detects this absence and reports an error. Similarly, if a robot attempts to place a plate when it is not currently holding one, this inconsistency triggers an error. The simulation also validates whether commands are executable given the robots' physical constraints, such as detecting when a requested position is outside a robot's reach or when joint limitations would prevent a specified movement. All errors are reported with a useful English-language description of the problem and any relevant details from the simulation, allowing the language model to more easily identify and reason over the problem. These runtime checks help identify issues that might not be apparent when manually reviewing a protocol but would cause failures during physical execution.

\subsubsection{Iteration Process}

When the simulation detects an error, we append the error as a new message in the conversation with the language model, which is sufficient for guiding the model to think about and fix the issue. Models exhibit different patterns when correcting errors in protocols, ranging from immediately identifying and fixing the issue, to misdiagnosing the problem and introducing ineffective solutions or even new errors. Through the use of the simulation, all of these possible behavior patterns are handled by iteratively prompting the model with detected errors from each attempt, allowing the model to explore and learn from its mistakes. This is done until the model converges to a protocol without any errors.

A successful result indicates that the physical aspects of the protocol should execute without issues in the real laboratory as the robots' movements, plate transfers, and device operations have all been verified in the digital twin. However, it's important to note that simulation success does not guarantee scientific accuracy. The simulation confirms physical executability but cannot verify that the protocol will achieve the intended scientific outcome, as we do not simulate liquid physics, chemical reactions, or biological processes. After a protocol passes simulation validation, it undergoes human review to verify scientific accuracy. In our work, a researcher still examines the final protocol to confirm that it performs the requested scientific operations in the correct sequence with appropriate parameters. This manual review step ensures that in the process of fixing physical execution issues the model did not introduced changes that would compromise the experiment's scientific validity.

\subsection{Real World Execution}\label{sec:Real World Execution}

% \ian{1) We assert that these two case studies are representative. Explain how.
% 2) We do not describe the two cases: we just say ``PCR'' and ``Cell Painting." Surely we need to provide an English language description of each, at least. 3) The descriptions that follow are incomplete and mix together descriptions of the protocol and the environment.
% 4) The terms case study, assay, benchmark are used, I am not sure what they are intended to convey.
% Both are case studies, but only PCR is a benchmark? Clarify why?
% How is Cell Painting an assay? That is not indicated that I can tell.
% 5) As I note below, it seems lame to me that Cell Painting was not executed. Shouldn't that be trivial to do?
% 6) The term ``Cell Painting" confuses me, because I think of the things being filled as Wells not Cells. But perhaps that is standard usage.
% }

We selected two representative experimental workflows to evaluate PRISM capabilities: a PCR amplification protocol and a Cell Painting assay. We choose these two workflows because they (i) encompass a range of laboratory operations, reagent types, and protocol structures; (ii) are widely used in biological research; (iii) are structurally distinct; and (iv) capture the types of reasoning challenges that PRISM is designed to address. PCR provides a compact, highly standardized molecular biology workflow, while Cell Painting represents a multi-step, multi-reagent imaging assay with substantially higher procedural complexity.

We selected \textbf{PCR amplification} as a first benchmark task because it is a canonical, well-defined molecular biology procedure with a clear, measurable outcome. Benchmarking against PCR allows direct comparison to prior work on protocol generation and provides a reproducible baseline for evaluating end-to-end execution on a robotic system.
In a typical PCR workflow, a reaction mixture containing a DNA template, primers, nucleotides, and polymerase is assembled and subjected to a series of temperature cycles to amplify the target DNA. Validation is commonly performed by gel electrophoresis. However, our automated laboratory does not include gel-handling capabilities. To enable fully autonomous execution, we used the Luna qPCR Master Mix (New England Biolabs), which incorporates an intercalating dye allowing real-time fluorescence readout. This adaptation provides an automated, quantitative signal equivalent to confirming the presence of an amplified product, without requiring any manual steps.

The protocol generated by PRISM was executed end-to-end on our automated laboratory platform, which integrates an Opentrons OT-2 liquid handler \cite{ot2}, a PF400 robotic arm \cite{pf400}, a Hidex plate reader \cite{hidex}, and an Azenta plate sealer \cite{sealer} and peeler \cite{peeler}. This experiment demonstrates that PRISM can produce robot-executable protocols that run to completion with no human intervention.

\begin{figure*}[t]
    \centering
    % adjust path/filename + width as needed
    \includegraphics[width=\textwidth]{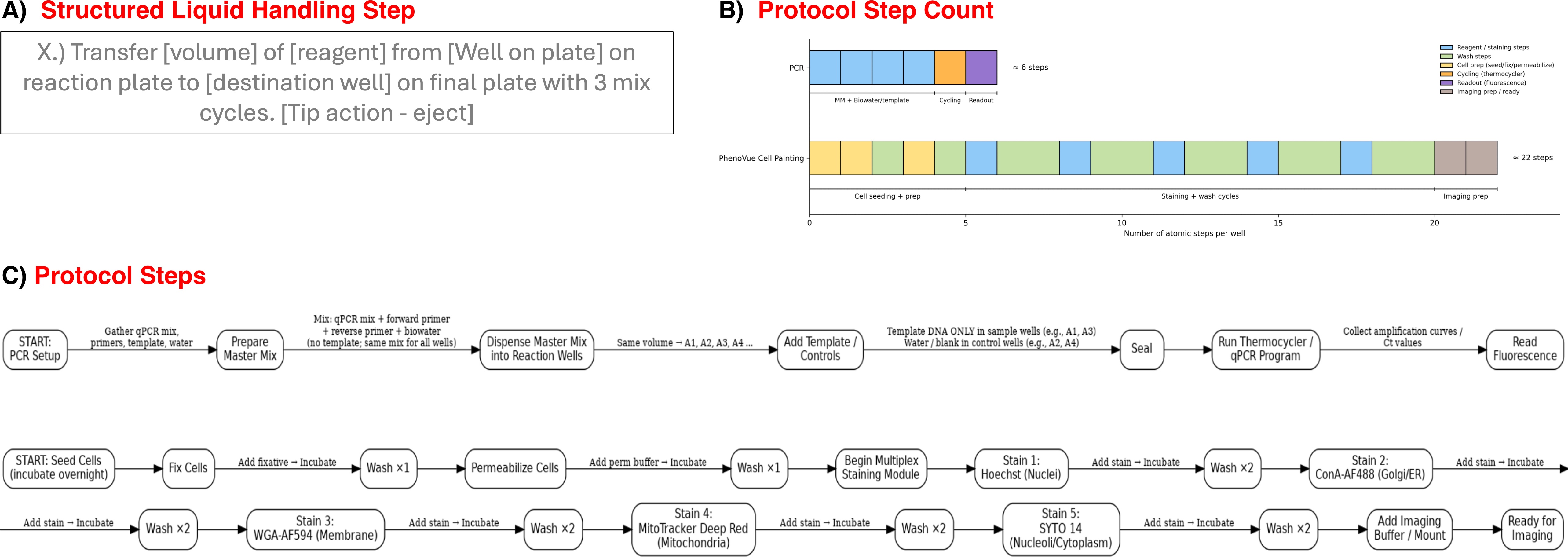}
    \caption{
        Comparison of protocol complexity for Luna qPCR and PhenoVue Cell Painting assays.
(A) Structured liquid-handling step format that all generated protocols must follow.
(B) Number of atomic steps required per well/reaction for each assay, coloured by
operation type. Cell Painting requires substantially more reagent additions and wash
cycles than qPCR, and is highly order-dependent (e.g., stain–wash cycles must be
performed sequentially), whereas reagent addition order does not matter in
qPCR. (C) Schematic step-by-step workflows for qPCR (top) and Cell Painting (bottom),
highlighting the increased length, repetition, and ordering constraints of the Cell
Painting protocol.
    }
    \label{fig:protocol-complexity}
\end{figure*}

To assess generality beyond PCR, we also applied PRISM to generate a protocol for a \textbf{Cell Painting assay}, a standard morphological profiling workflow widely used in high-content imaging screens. In a typical Cell Painting assay (see Figure~\ref{fig:protocol-complexity}), adherent cells are seeded, fixed, permeabilized, and stained using a panel of fluorescent dyes that highlight cellular structures such as nuclei, endoplasmic reticulum, Golgi, and mitochondria. The stained cells are then imaged using a high-content screening microscope to extract morphological features.
Cell Painting exercises protocol-generation capabilities that PCR does not: multi-step washing and incubation, sequence-dependent operations, complex reagent handling, and long branching workflows. 
These characteristics make it an ideal test of PRISM’s ability to handle procedurally complex, multi-module protocols.

We evaluated the Cell Painting protocol produced by PRISM in silico. A domain expert reviewed the generated steps for biological correctness and internal consistency. Although the core Cell Painting steps (fixation, permeabilization, staining, and washing) can be emulated on our OT-2 through controlled pipetting, our current workcell does not include an integrated automated plate washer. Because Cell Painting typically involves numerous gentle wash cycles optimized for preserving adherent cell monolayers, dedicated plate washers are commonly used in high-content screening workflows. For this reason and because our platform lacks a high-content imaging system, we validated the Cell Painting protocol in silico rather than executing it physically.
% To demonstrate the capabilities of PRISM, we selected two representative case studies (PCR and Cell Painting). PCR amplification was used for experimental validation in our automated laboratory setup, while a Cell Painting assay was used to test in silico protocol generation. Together, these cases illustrate both the feasibility of end-to-end execution and the generalizability of PRISM to more complex assays.

% We chose PCR as a benchmark because it is a standard, well-defined molecular biology task. While PCR products are typically validated by gel electrophoresis, our laboratory does not include gel-handling automation. To enable end-to-end execution without manual intervention, we used the Luna qPCR master mix (New England Biolabs), which provides real-time fluorescence readout as a substitute for gel analysis. This adaptation allowed us to confirm experimental success directly from robotic execution, demonstrating that PRISM can produce fully automated, executable wet-lab protocols. The protocols generated by PRISM were executed on our automated laboratory platform, consisting of an Opentrons OT-2 liquid handler \cite{ot2}, a PF400 robotic arm \cite{pf400}, a Hidex plate reader \cite{hidex}, and an Azenta sealer \cite{sealer} and peeler \cite{peeler}.  

% To assess generality beyond PCR, we also applied PRISM to generate a Cell Painting assay. The Cell Painting pipeline was validated in silico and examined by a biologist for its experimental correctness but was not validated on the robotic platform.

%% file: sections/results.tex
\section{Results and Discussion}

We present our results in an order that reflects the three-stage pipeline of the PRISM framework illustrated in Figure~\ref{fig:prism_overview}. Section~\ref{sec:Protocol Planning} evaluates Stage 1, where we assess different approaches to protocol planning that convert high-level experimental intent into structured English instructions. This includes comparing multi-agent versus single-agent frameworks and constrained versus open-ended prompting strategies. Section~\ref{sec:Protocol Generation / Validation} corresponds to Stage 2, the protocol generation and validation loop, where structured steps are translated into robot-executable YAML formats and iteratively refined through simulation-based error detection. We also present an ablation study demonstrating the necessity of simulation for achieving physical correctness. Finally, Section~\ref{sec:Real World Experimental Validation} presents Stage 3, where simulation-validated protocols are executed on our automated laboratory platform to demonstrate end-to-end functionality and biological validity.

\begin{figure*}[t]
  \centering
  \begin{subfigure}{0.49\textwidth}
    \includegraphics[width=\linewidth,keepaspectratio]{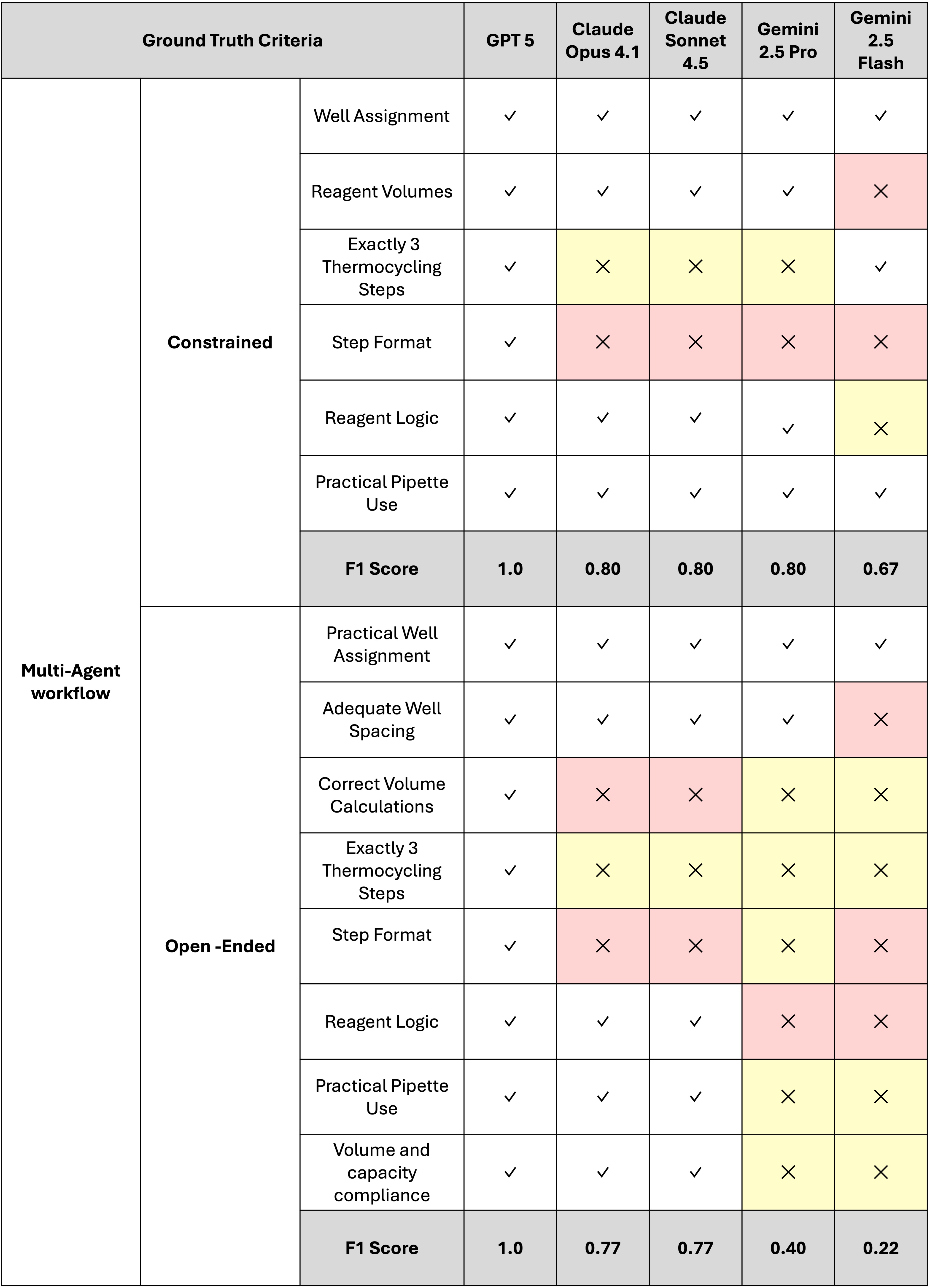}
    \caption{Multi-agent}
    \label{fig:multi-agent-sub}
  \end{subfigure}\hfill
  \begin{subfigure}{0.49\textwidth}
    \includegraphics[width=\linewidth,keepaspectratio]{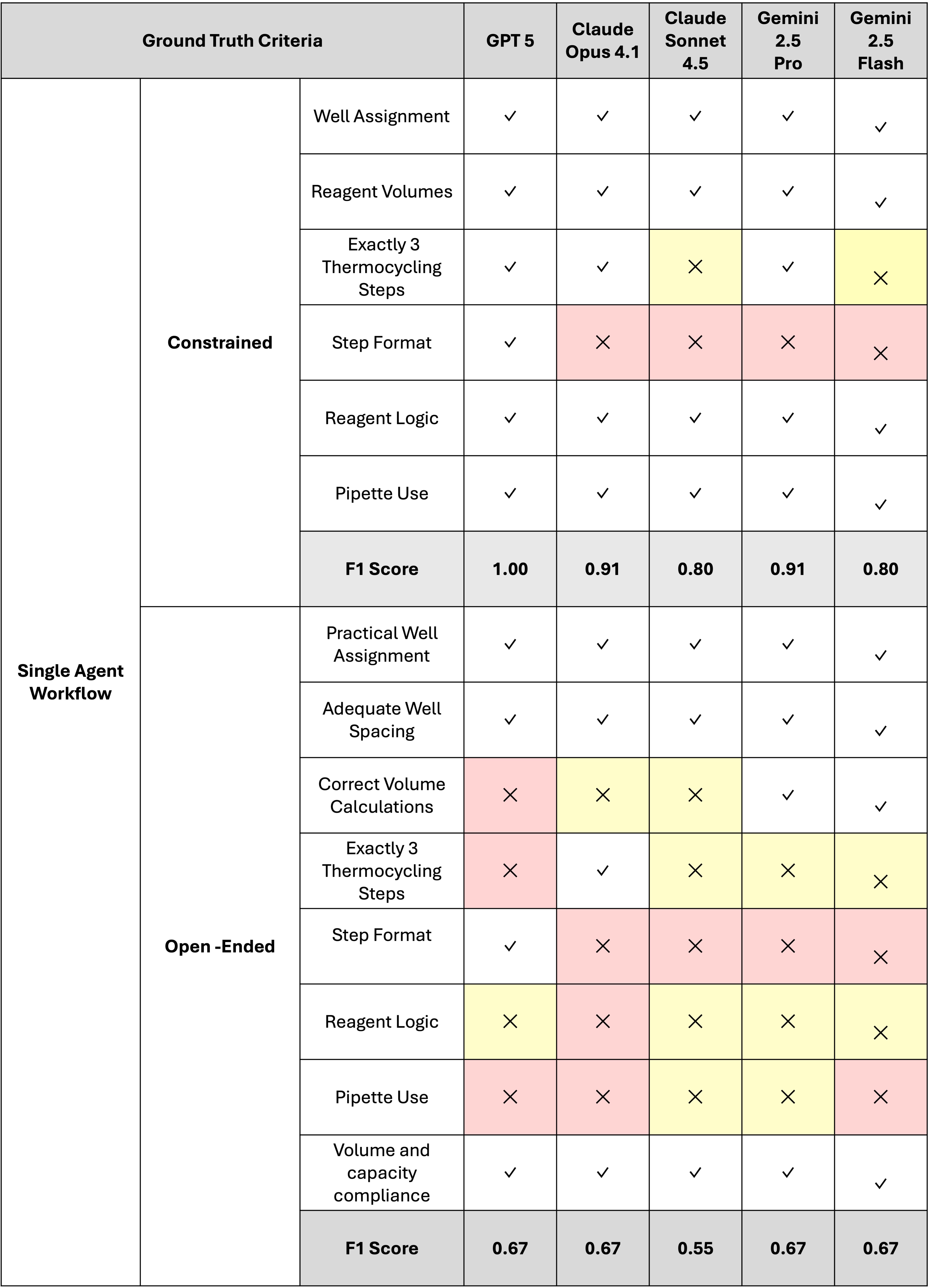}
    \caption{Single-agent}
    \label{fig:single-agent-sub}
  \end{subfigure}
  \caption{\textbf{Comparison of Multi-Agent and Single-agent Workflow Performance on Luna qPCR Protocol Generation.}
  Each panel summarizes per-model accuracy across all ground-truth evaluation criteria for five LLMs: GPT-5, Claude Opus 4.1, Claude Sonnet 4.5, Gemini 2.5 Pro, and Gemini 2.5 Flash. \checkmark\ and \xmark\ indicate whether each criterion was met or violated. Color coding denotes the type of deviation from the ground truth: yellow = extra action (false positive), red = missing action (false negative). These errors contribute to the F1-score reported in the bottom row, which quantifies overall protocol accuracy.
Panel (a) shows the multi-agent workflow (WebSurfer → Planner → Critique → Validator), and Panel (b) shows a single agent workflow without modular decomposition of tasks.
The comparison highlights that structured multi-agent reasoning achieves higher correctness and more reliable protocol synthesis than monolithic reasoning generation.}
  \label{fig:model-evals}
\end{figure*}

\subsection{Protocol planning} \label{sec:Protocol Planning}

We begin our evaluation by focusing on the initial phase of the PRISM framework: the translation of scientific intent into structured experimental plans. This stage serves as the foundational logic layer; errors introduced here propagate downstream, rendering even syntactically perfect robotic code scientifically invalid. To assess this capability, we benchmark performance across two dimensions: the architectural framework (comparing single-agent reasoning against the proposed multi-agent system) and the prompt constraints (evaluating adaptability under both rigid and open-ended instructions).

\subsubsection{Experimental Setup and Evaluation Criteria}

We evaluated five LLMs(GPT-5, Claude Opus 4.1, Claude Sonnet 4.5, Gemini 2.5 Pro, and Gemini 2.5 Flash) on their ability to generate structured liquid handling steps in plain English for the PCR and Cell Painting workflows described in Section~\ref{sec:Real World Execution}. 

For PCR, each model was tested under both the multi-agent and single-agent frameworks using both the constrained and open-ended prompting paradigms. Performance was assessed based on (1) the number of correction iterations required to refine the generated protocol and (2) the types of logical, formatting, or physical errors present in the generated steps. We allowed at most three correction iterations; if a model failed to produce a valid protocol within this limit, the attempt was recorded as non-convergent. The PCR results were reviewed by a biologist and experimentally validated on our robotic platform. 

We evaluated each generated protocol for both structural correctness and biological feasibility. 
Under the \textbf{constrained prompting} setup, models received fixed experimental parameters, reagent mappings, and well layouts designed to minimize ambiguity, ensuring that success reflected precise adherence to given instructions rather than creative inference. Performance was assessed across six predefined criteria: (1) correct source and destination well assignments, (2) correct reagent volumes, (3) inclusion of exactly three thermocycling steps, (4) adherence to the required step format, (5) correct reagent use for test and control wells, and (6) appropriate pipette selection. 

In contrast,  the \textbf{open-ended prompting} setup provided only high-level experimental goals without fixed reagent mappings or per-component volumes, while still specifying essential lab constraints such as the OT-2 deck layout (e.g., slot assignments for the reaction and destination plates) and a target total reaction volume. Therefore, models were  required to infer compatible volume splits, well layouts, and transfer sequences that satisfied these physical and biological constraints. Evaluation in this setting focused on eight criteria that capture reasoning flexibility and practical feasibility: (1) practical well mapping, (2) adequate spacing between test and control wells to avoid cross-contamination, (3) biologically correct volume calculations, (4) inclusion of exactly three thermocycling steps, (5) correct step formatting, (6) correct reagent logic for test and control wells, (7) proper and consistent pipette use, and (8) compliance with minimum volume and plate capacity limits. 

% Compact cycles table (models × {Multi-Agent, Reasoning} × {Constrained, Open-ended})
\begin{table}[t]
\caption{Luna qPCR Protocol results showing the number of refinement iterations (1–3) required for each model to achieve convergence under the multi-agent and single-agent frameworks across constrained and open-ended prompting. \checkmark\ indicates that the model produced a correct protocol on the first attempt without requiring refinement; integers (1–3) denote the number of refinement cycles needed before convergence; and \dag\ indicates non-convergence within three iterations.}
\label{tab:cycles-compact}
\centering
\resizebox{\columnwidth}{!}{%
\begin{tabular}{lcccc}
\toprule
& \multicolumn{2}{c}{\textbf{Multi-Agent}} & \multicolumn{2}{c}{\textbf{Single-agent}} \\
\cmidrule(lr){2-3}\cmidrule(lr){4-5}
\textbf{Model} & \textbf{Constrained} & \textbf{Open-ended} & \textbf{Constrained} & \textbf{Open-ended} \\
\midrule
GPT-5 & \checkmark & \checkmark & \checkmark & 2 \\
%  & \checkmark & 2 & \checkmark & 2 \\
Claude Opus 4.1 & 2 & \dag & \dag & \dag \\
Claude Sonnet 4.5 & \dag & \dag & \dag & \dag \\
Gemini 2.5 Pro & \dag & \dag & \dag & \dag \\
Gemini 2.5 Flash & \dag & \dag & \dag & \dag \\
\bottomrule
\end{tabular}}
\end{table}

We provide the full prompt templates and model instructions for both the constrained and open-ended paradigms in \autoref{sec:Appendix Protocol Planning}. Quantitative accuracy across the criteria listed above was measured using the F1 score, which captures both the precision and completeness of the generated protocol relative to the ground-truth reference. The F1 score was calculated as,
\[
F_1 = \frac{2 \times \text{TP}}{2 \times \text{TP} + \text{FP} + \text{FN}}
\]
where TP, FP, and FN represent the number of correct, additional, and missing steps, respectively. The F1 score accounts for additions, deletions, and errors in various ground truth criteria chosen for each prompt paradigms.

\subsubsection{Use Case 1: Visual Exploration of PRISM Capabilities}
\begin{figure*}[t]
    \centering
    \includegraphics[width=\textwidth]{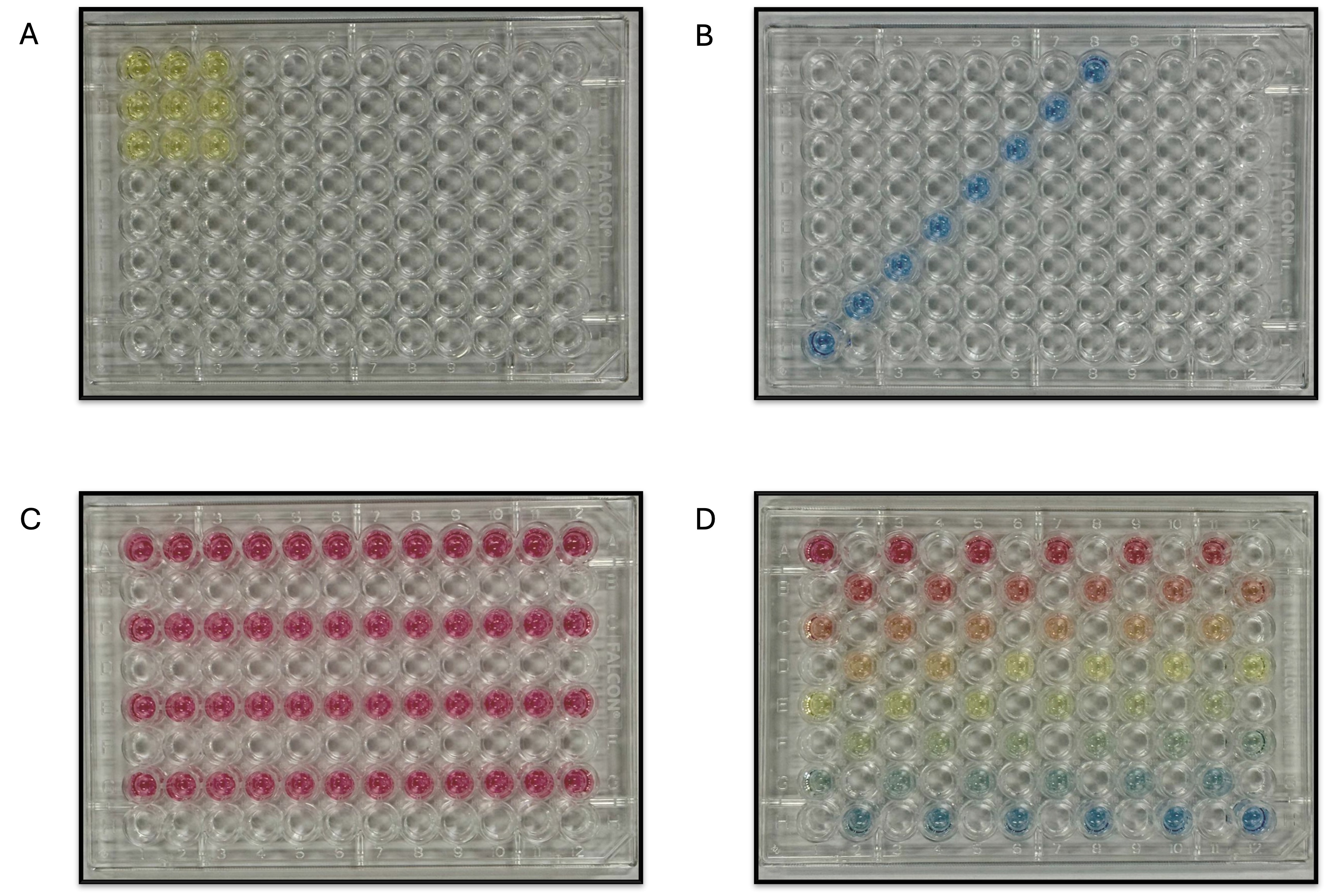}
    \caption{
        \textbf{Visualization-focused pattern-generation tasks used to evaluate PRISM.}
        (A) A $3 \times 3$ coloured square in the upper-left corner. 
        (B) A diagonal pattern beginning at the lower-left corner (H1). 
        (C) An alternating-row fill pattern. 
        (D) A rainbow checkerboard combining a red–yellow–blue gradient with a parity-based fill rule.
    }
    \label{fig:prism-toy-patterns}
\end{figure*}

We first evaluated PRISM in our Rapid Prototyping Laboratory (RPL) on a set of plate-pattern tasks to demonstrate its ability to generate correct OT-2 protocols outside a biological laboratory. Specifically, we validated PRISM's spatial planning capabilities by replicating the geometric pattern tasks previously demonstrated by Boiko et al.~\cite{boiko2023emergent}, confirming that our system can successfully interpret high-level visual descriptions into robot-executable protocols. These non-biological demonstrations show that PRISM can infer well locations, compute volumes, assign reagent sources, and output robot-executable liquid-handling instructions based solely on a natural-language goal.
Figure \ref{fig:prism-toy-patterns} illustrates the four patterns used: (A) a 3 $\times$ 3 colored block in the upper-left corner, (B) a diagonal starting from the lower-left corner of the plate, (C) an alternating-row fill pattern, and (D) a rainbow checkerboard created by applying a color gradient to every other well. These tasks highlight PRISM’s capacity to generalize beyond assay-specific workflows and generate protocols for arbitrary user-defined objectives. Using only an English-language description of which wells to fill and how to color each well, PRISM determines exact well locations and ratios of input liquids to complete the task.

\subsubsection{Use Case 2: PCR Protocol Generation} \label{sec:Use Case 2}

Next, we evaluated PRISM in RAPID-446, our BSL-1 laboratory, to show that PRISM generalizes to different laboratory configurations and different instruments. RAPID-446 contains additional robots in a different physical layout to that of RPL, allowing us to demonstrate that PRISM is not restricted to the RPL configuration that we used for development. 

Figures~\ref{fig:multi-agent-sub} and~\ref{fig:single-agent-sub} summarize per-model F1 accuracy for PCR across the two frameworks, while Table~\ref{tab:cycles-compact} reports the number of correction iterations required for each model to converge. Together, these analyses capture both the accuracy of protocol generation and the adaptability of each framework under iterative refinement.

Under constrained prompting, GPT-5 achieved perfect F1 scores for both the multi-agent and single-agent frameworks, generating a correct protocol in the first attempt. 
Under open-ended prompting, GPT-5 initially showed minor inconsistencies in volume calculations, step formatting, number of thermocycling steps, and transfer of reagents but converged to fully correct protocols within three refinement iterations. 

In contrast, the Claude and Gemini models did not converge reliably across iterations, with several configurations showing degraded performance over time. For example, Gemini 2.5 Flash (reasoning with constrained prompts), Claude Opus 4.1 (reasoning with fixed constraints), and Claude Opus 4.1 (open-ended multi-agent framework) often added redundant steps, scaled volumes to impractical levels, or altered previously correct actions. These models also struggled to maintain the required step format and failed to produce exactly three thermocycling steps under constrained prompting. 
When open-ended prompts were used they deviated further across evaluation criteria particularly in maintaining biologically valid volume or concentration calculations compatible with the robotic platform. Overall, the multi-agent framework still achieved higher accuracy and faster convergence for most non-GPT models, largely because its structured feedback helped recover from format and logic inconsistencies that single reasoning chains could not self-correct. However, they still had minor issues in step-format and therefore we say they did not fully converge in Table~\ref{tab:cycles-compact}. 

These results indicate that model performance is shaped not only by parameter capacity but also by reasoning dynamics: the GPT model exhibits sustained contextual reasoning that allows it to manage experimental constraints implicitly, whereas models like Gemini and Claude, which favor faster response generation, often lose context or respond without sufficient internal deliberation, requiring more explicit structure and corrective feedback to achieve valid protocols.

\subsubsection{Use Case 3: Cell Painting} \label{sec:Use Case 3}

% \ian{1) 
% Do we show the prompt for this?}

To evaluate performance on more complex experimental workflows, we applied both the multi-agent and single-agent frameworks to generate a Cell Painting protocol. This evaluation was not benchmarked across prompt paradigms, as the goal was to test whether each framework could handle a longer, multi-step procedure rather than optimize specific prompting conditions. The generated protocols were reviewed by a biologist to confirm that they were biologically sound and physically feasible on our automated platform. In this setting, reasoning models often failed to maintain coherent step structure as the number of actions increased, frequently omitting reagents or misordering procedural steps. In contrast, the multi-agent framework produced a consistent and logically organized protocol that captured the necessary sequence of reagent additions and washing steps. The division of tasks across specialized agents helped maintain structure and correctness as procedural complexity increased, underscoring the importance of modular, feedback-driven design for planning larger, multi-phase experiments.

Overall, these results show that structured prompting and modular task decomposition play complementary roles in generating correct and physically executable experimental protocols. High-capacity models such as GPT-5 perform reliably across both frameworks and prompt types, while smaller models benefit substantially from the structured feedback of the multi-agent system and the explicit guidance of constrained prompts. The multi-agent framework offers a clear advantage when protocols become longer or more detailed, enabling consistent correction of logical and formatting errors that reasoning models alone struggle to resolve. Together, these findings demonstrate accurate and adaptable protocol planning and provides a foundation for the next stage of translating the structural steps into fully executable robotic workflows.

\subsection{Protocol Generation / Validation}\label{sec:Protocol Generation / Validation}

\begin{figure}[ht]
    \centering
    \includegraphics[width=\columnwidth]{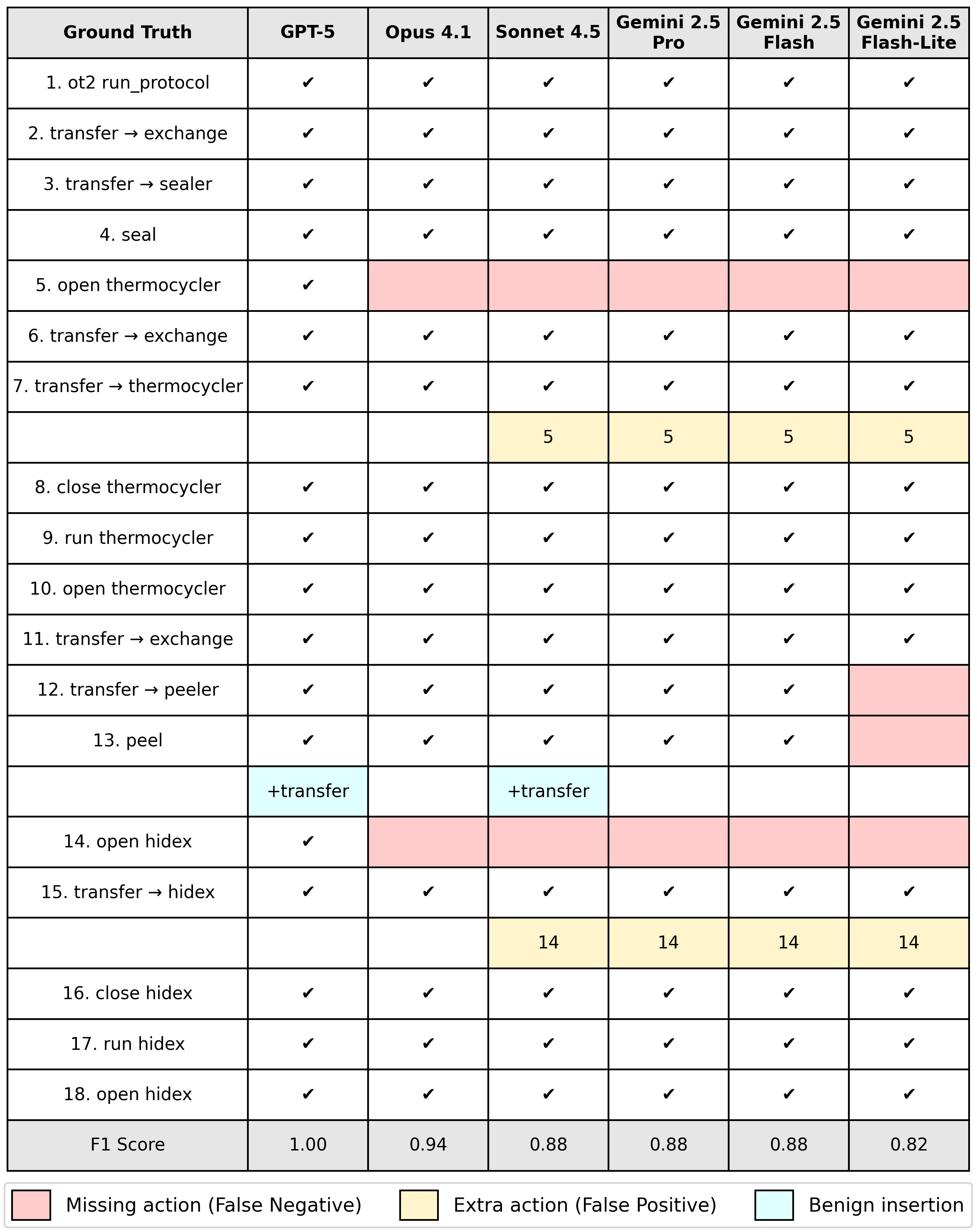}
    \caption{Initial generated protocols vs.\ ground truth for a PCR workflow. Each column shows the first protocol generated by the specified reasoning model compared against the correct sequence of actions. \checkmark\ indicates correct action placement. Red cells indicate missing actions, yellow cells indicate inserted actions (the number corresponds to correct position of the action), and blue cells indicate benign modifications where the model added unnecessary but non-harmful intermediate steps. GPT-5 achieved perfect initial generation, while other models exhibited a common failure pattern: omitting the \texttt{open} commands required before transferring plates into the thermocycler (step~5) and plate reader (step~14).}
    \label{fig:initial_protocols}
\end{figure}

To evaluate the initial protocol generation capabilities of each model, we assessed their ability to produce a complete, executable PCR workflow from structured English instructions in a single attempt. This zero-shot evaluation provides a baseline measure of each model's understanding of robotic constraints, sequencing requirements, and the physical limitations of the laboratory environment. Figure~\ref{fig:initial_protocols} presents a comparative analysis of the first protocol generated by each of the six tested reasoning models against the ground truth workflow.

The analysis reveals a striking pattern across the models: five of the six tested systems (Claude Opus 4.1, Claude Sonnet 4.5, Gemini 2.5 Pro, Gemini 2.5 Flash, and Gemini 2.5 Flash-Lite) all omitted the \texttt{open} commands required before placing plates into the thermocycler and plate reader. This error type represents a physical impossibility: attempting to insert a plate into a closed device would cause a failed experiment and potential equipment damage in the physical laboratory. The error pattern demonstrates a consistent gap in long-horizon planning: while models successfully generated most of the basic protocol structure, they failed to anticipate that certain devices require extra operations before interaction, a constraint that differs from the typical usage pattern of the other laboratory robots.

Notably, GPT-5 achieved perfect initial generation with an F1 score of 1.0, producing a physically executable protocol without any errors on the first attempt. The performance gradient across models, reflected in declining F1 scores from GPT-5 (1.0) through Claude Opus 4.1 (0.94) down to Gemini 2.5 Flash-Lite (0.82), shows quantitatively how model capability directly impacts initial generation quality. Less capable models exhibited higher rates of missing actions and a failure to fully capture the basic protocol structure.

These zero-shot results establish the fundamental challenge that the PRISM framework addresses: even state-of-the-art language models often produce physically infeasible protocols when translating high-level scientific intent into executable robotic instructions. The errors shown in Figure~\ref{fig:initial_protocols} would cause failures if executed directly on physical hardware, motivating the core function of PRISM's simulation-based refinement loop to iteratively detect and correct these flaws before real-world execution.

\subsubsection{Iterative Refinement Through Simulation Feedback}

All models except GPT-5 required further iterations with simulation feedback to produce valid protocols. The simulation environment detected physical impossibilities, such as attempting to place plates into closed devices, and returned error messages describing the specific collision or constraint violation. Models demonstrated the ability to interpret these error messages, identify the problematic steps, and propose corrections, though the efficiency and pattern of error correction varied by model. %capability.

\begin{table}[ht]
\caption{Number of simulation iterations required for each model to produce a physically valid PCR protocol in simulation. GPT-5 generated a correct protocol without any errors in the first simulation run, while weaker models required progressively more attempts to identify and correct all errors. Gemini 2.5 Flash-Lite passed all simulation checks in 3 iterations, however it failed to include the \texttt{peel} command, a scientifically necessary step for correct results from the plate reader that our physics simulation did not verify.}
\label{tab:iteration_requirements}
\centering
\begin{tabular}{lc}
\hline
\textbf{Model} & \textbf{Total Iterations} \\
\hline
GPT-5 & 1 \\
Claude Opus 4.1 & 2 \\
Gemini 2.5 Pro & 2 \\
Claude Sonnet 4.5 & 3 \\
Gemini 2.5 Flash & 3 \\
Gemini 2.5 Flash-Lite & 3 (failed scientific requirements) \\
\hline
\end{tabular}
\end{table}

\begin{figure}[bht]
    \centering
    \includegraphics[width=0.7\linewidth]{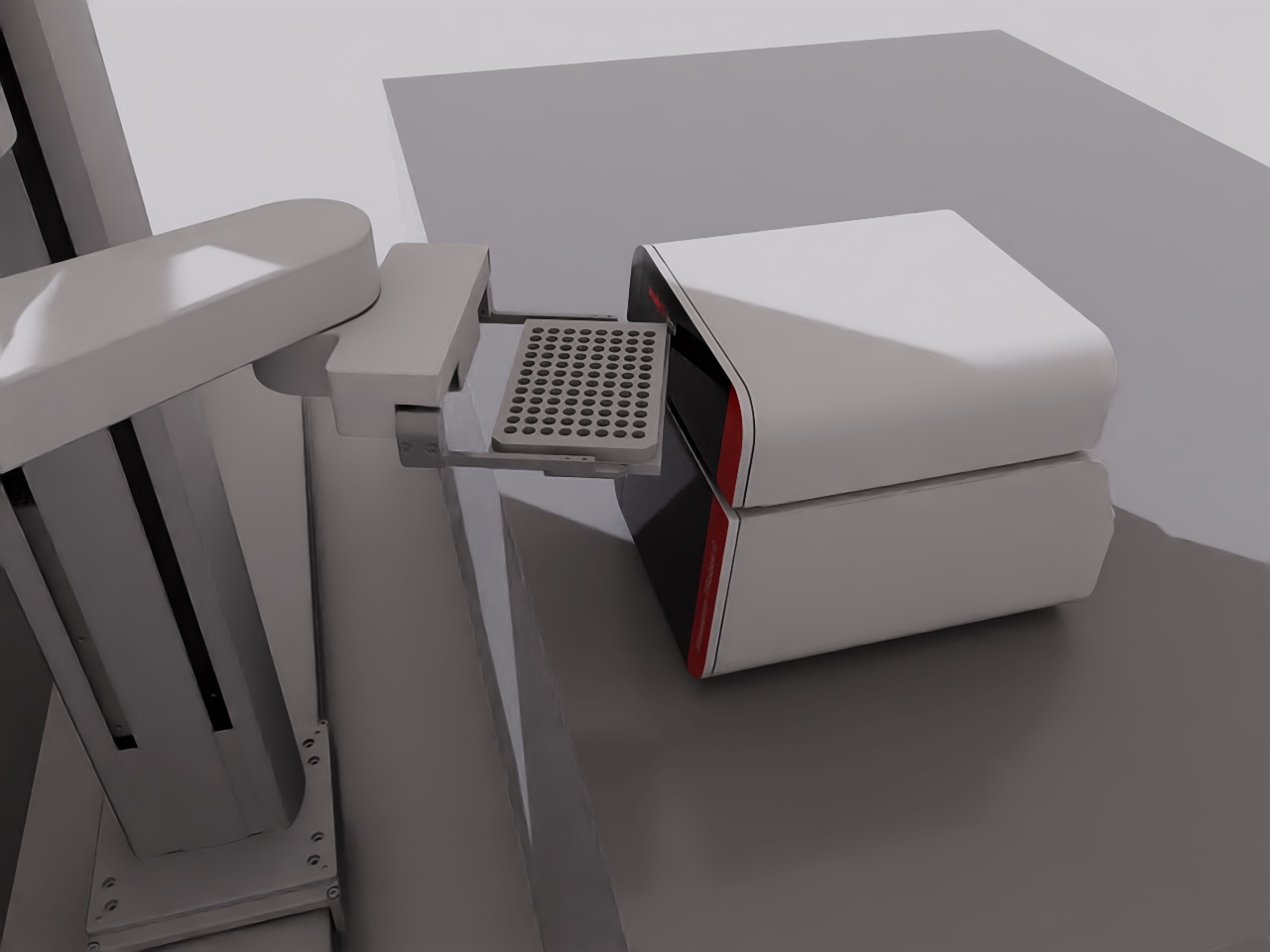}
    \caption{Visualization of a plate collision error in the simulation environment. The robot arm attempts to insert a plate into the thermocycler while the device lid remains closed, representing a physical impossibility. This error occurred when models omitted the required \texttt{open} command before plate insertion.}
    \label{fig:device_not_opened_error}
\end{figure}

The medium-capability models, Claude Sonnet 4.5 and Gemini 2.5 Flash, successfully interpreted simulation error messages and correctly diagnosed the root causes of failures. However, these models exhibited a localized correction pattern: when the simulation flagged a specific instance of an error (e.g., attempting to place a plate into a closed device at a specific step, as shown in Figure~\ref{fig:device_not_opened_error}), the model would fix only that particular occurrence. Subsequent simulation runs would then generate new error messages for other instances where the same mistake had been made, requiring additional iterations to systematically eliminate all occurrences of the error pattern. This behavior resulted in three iterations for both Claude Sonnet 4.5 and Gemini 2.5 Flash to achieve valid protocols, as shown in Table~\ref{tab:iteration_requirements}.

In contrast, the stronger models, Claude Opus 4.1 and Gemini 2.5 Pro, demonstrated global pattern recognition. When provided with an error message about a single instance of a mistake, these models recognized that they had made the same error in multiple locations throughout the protocol and corrected all instances simultaneously in a single iteration. This capability to generalize from one error instance to identify and fix a systematic problem pattern enabled these models to achieve valid protocols in just two iterations.

The weakest tested model, Gemini 2.5 Flash-Lite, exhibited substantially less capable error diagnosis behavior. Rather than immediately identifying the correct root cause, this model incorrectly handled the initial error and introduced new errors in the process. Only after accumulating additional context from further error messages did the model converge on a physically possible solution. This trial-and-error approach resulted in three iterations before achieving a solution that resolved the physics-based errors detected by the simulation. However, despite producing a physically executable protocol that passed all simulation checks, the model still failed to include the \texttt{peel} command, a scientifically necessary step for correct results from the plate reader that our physics simulation did not verify. When the human researcher provided a manually formatted error message indicating the missing operation, Gemini 2.5 Flash-Lite successfully incorporated a correction in its final iteration. This complete refinement process resulted in four total iterations before achieving both a physically valid and scientifically complete protocol, demonstrating that while simulation-based feedback substantially aids weaker models in achieving physical correctness, it does not eliminate the need for scientific review of the final protocol.

To assess the generalizability of these results beyond the PCR experiment, we also evaluated all models on a Cell Painting protocol. In this case, every model produced a correct protocol on the first attempt, requiring zero iterations for validation. This universal success was due to the fact that the majority of the complexity in the Cell Painting protocol lies in the liquid handling steps covered thoroughly by the planning phase (Figure~\ref{fig:prism_overview}, Stage 1) as shown in Figure~\ref{fig:protocol-complexity}, as well as the reduced robotic variety of the Cell Painting workflow, which lacked the intricate coordination requirements between robots that induced most of the model errors with the PCR protocol. Despite this universal first-attempt success, minor differences in model behavior remained observable. Each model independently selected its own interpretation of room temperature, with values varying between 22 and 25 degrees Celsius across different models. Additionally, when prompted to seal a plate at 4 degrees Celsius for an indefinite storage period, all models except GPT-5 chose to seal the plate without placing it in the incubator, leaving a note for the human researcher to configure storage as desired. This approach reflected a conservative interpretation of the input documentation, which did not describe indefinite-duration incubation as a supported capability. GPT-5 opted to place the sealed plate in the incubator at the required temperature for a specified duration of 24 hours, accompanied by a note for the researcher to adjust the duration as needed.

\begin{figure}[ht]
    \centering
    \includegraphics[width=0.7\linewidth]{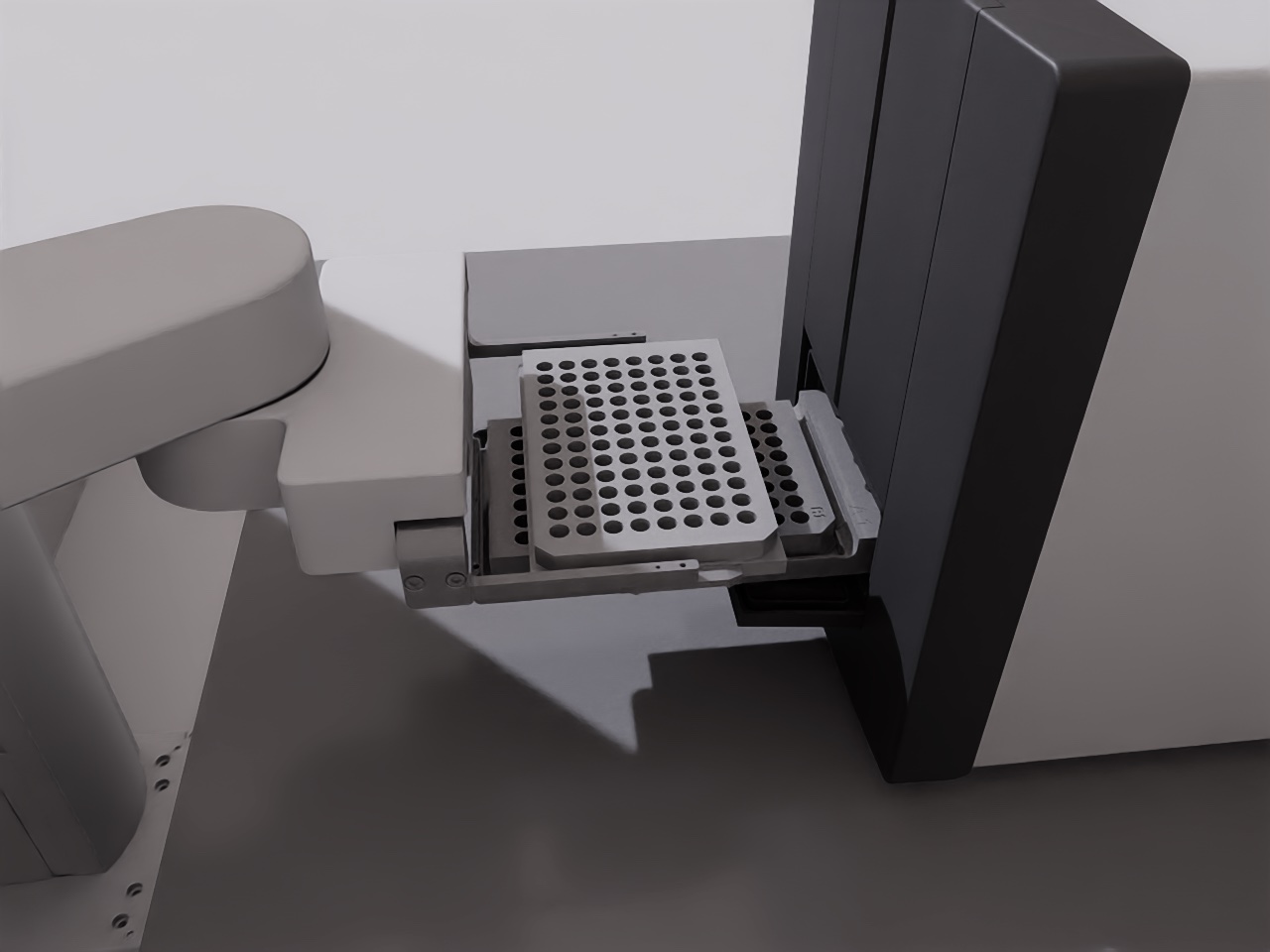}
    \caption{Visualization of a plate orientation mismatch error in the simulation environment. The plate is incorrectly oriented during a transfer operation, resulting in a spatial configuration that would cause a collision with the robot arm or device during execution. Despite multiple rounds of self-correction where the model declared the protocol ready for execution, this type of subtle geometric constraint violation remained undetected, demonstrating the limitations of purely text-based reasoning for validating physical feasibility.}
    \label{fig:orientation_error}
\end{figure}

\subsubsection{Limitations of Self-Correction Without Simulation} % An Ablation Study}

To evaluate the value that simulation-based validation provides beyond language model self-critique, we conducted an ablation study where the simulation environment was disabled. This experiment tested whether a reasoning model, when prompted to identify and fix its own errors, could achieve the same level of physical correctness as the complete PRISM framework with simulation feedback. For this work, we designed a more challenging scenario to stress-test the model's abstract reasoning capabilities: a protocol that processes two plates sequentially through the PCR workflow, creating a conflict where both plates would need to occupy the plate reader simultaneously, a physical impossibility. This scenario simulates a realistic failure mode where the upstream protocol planning phase produces a logically coherent but physically infeasible plan.

We provided Claude Sonnet 4.5 with this flawed plan and asked it to generate the corresponding YAML protocol. The initial generation contained multiple errors: the model forgot to include \texttt{open} commands for both the thermocycler and plate reader before attempting to insert plates, and it produced mismatched plate transfer orientations that would cause collisions. Additionally, the protocol's terminal state had both plates in the plate reader simultaneously, confirming that the model had not detected the physical impossibility during generation.

When prompted to review its work and identify any mistakes, the model successfully recognized the high-level logical conflict: two plates could not occupy the same location. It proposed a solution to move the first plate to an exchange deck before processing the second plate, demonstrating capable logical reasoning about spatial conflicts. However, this modification introduced a new problem. The protocol required use of the exchange deck as an intermediate transfer location for the second plate's movements, and with the first plate now occupying this position, the exchange deck was blocked. Upon a second self-critique request, the model recognized this emergent conflict and revised the plan again, orchestrating movements of the first plate back to its original position in the OT-2 to free both the plate reader and the exchange deck.

After this sequence of corrections, the model performed a final self-review and declared the protocol to be correct and ready for execution. However, multiple physical impossibilities remained undetected. The protocol still attempted to place plates into closed devices, the same error pattern shown in Figure~\ref{fig:initial_protocols} that all models except GPT-5 produced in their initial generations. Additionally, the mismatched plate transfer orientations persisted (Figure~\ref{fig:orientation_error}), which would cause collision failures during execution.

In the complete PRISM framework, this protocol that the model declared ``correct'' would have failed when executed in the simulation environment. The grounded, physics-based error messages provide a fundamentally different form of validation than self-critique, enabling the model to identify and correct errors that it could not detect through tool-unassisted reasoning.
%The simulation acts as a mandatory validation gate, catching the subtle physical infeasibilities that can slip through language-based verification and preventing protocols with ``silent failures'' from reaching the physical laboratory where they would cause equipment damage, material waste, or experimental failure.

This ablation study reveals the limitations of language model self-correction for protocol refinement. The model demonstrated some sophisticated logical reasoning, successfully identifying and resolving high-level spatial conflicts through multiple iterations of abstract problem-solving. However, given the complexity of the multi-step coordination task and the subtle nature of certain physical constraints, text-based reasoning alone was insufficient to achieve a physically executable outcome.

% \ian{I like this discussion but also wonder how much you can generalize from this example. Indeed you showed that there was an error that simulation detected but the model did not. You write in absolutist terms, implying that this error had some fundamental property that prevented detection by what you call ``abstract reasoning'' (by which I think you mean LLMs)--making the simulation an ``irreplaceable validation layer.'' If so, can you be more formal about the nature of the errors that can only be detected in this way? Or could it be detected by a different model or different prompt? A more reserved take on things would be to say simply that you have found that simulation can increase confidence by catching errors that this particular LLM, with this particular prompt, did not. You can't guarantee, I don't think, that the simulation will catch all errors, or that an LLM could not capture this particular error? Two related questions: a) Are there errors that simulation cannot catch, but reasoning can? b) Are there errors that neither can capture reliably? (e.g., race conditions?)}

This result establishes that the simulation component of PRISM provides an valuable validation layer. While language models possess remarkable reasoning capabilities and can identify many types of errors through self-critique, the combination of task complexity and the need to verify physical feasibility requires grounding in a physics-based environment. The simulation complements rather than replaces model reasoning, providing the physical validation that tool-unassisted reasoning cannot reliably achieve for complex, multi-step robotic protocols.

\subsection{Real World Experimental Validation}\label{sec:Real World Experimental Validation}

\begin{figure}[ht]
    \centering
    \fbox{\includegraphics[width=0.8\linewidth]{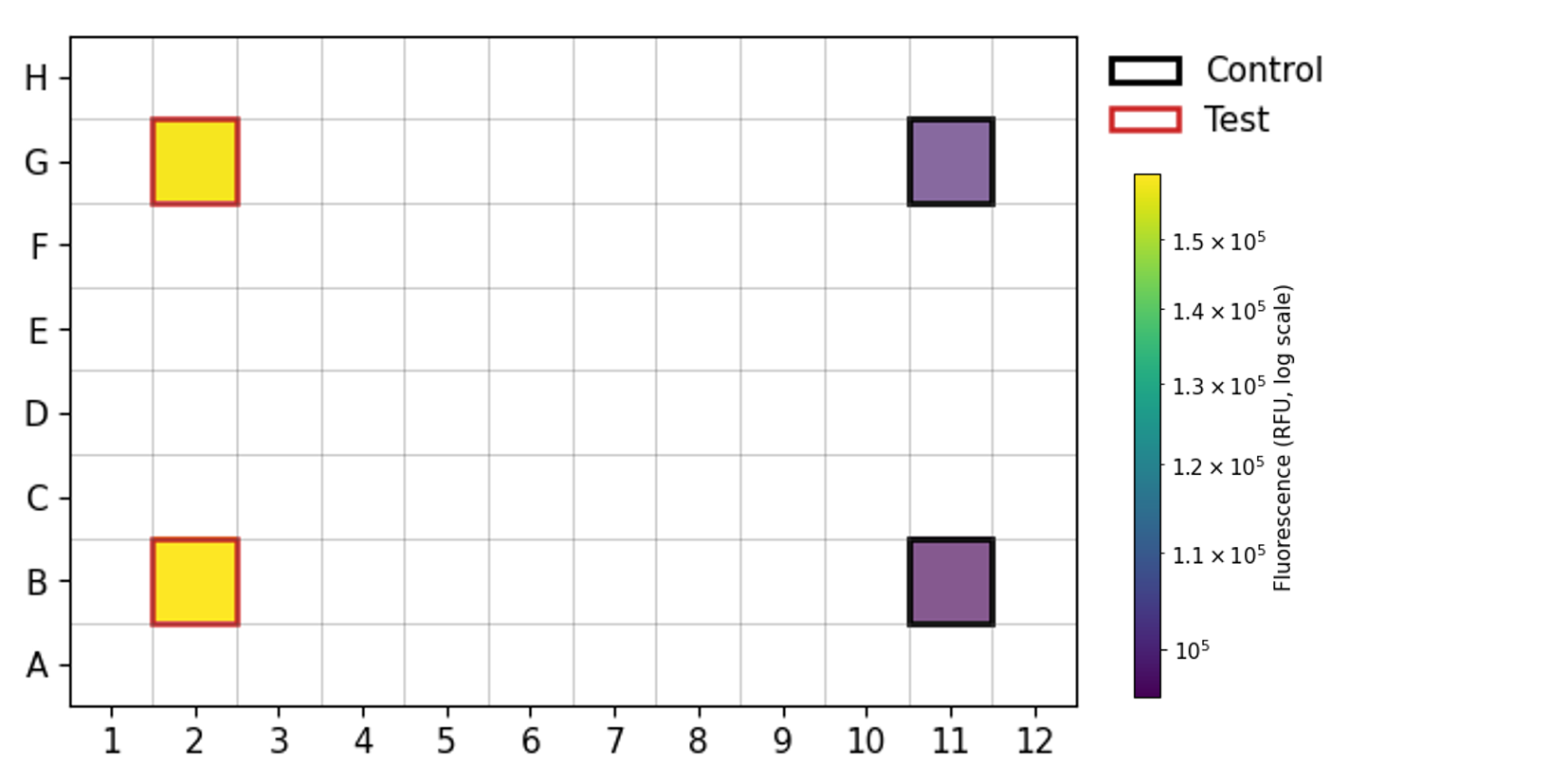}}
    \caption{PCR plate fluorescence readout from PRISM-executed protocol. Test wells (red outline) show amplified signal, while control wells (black outline) remain at baseline. Fluorescence is shown in relative fluorescence units (RFU) on a log scale.
    }
    \label{fig:pcr_validation}
\end{figure}

We executed a PCR protocol generated by PRISM, as described in Section~\ref{sec:Use Case 2}, on our automated laboratory platform. We used the Luna qPCR master mix, which enables fluorescence-based validation in place of gel electrophoresis, allowing fully automated execution. As shown in Figure~\ref{fig:pcr_validation}, test wells (outlined in red) produced strong amplification signals, while control wells (outlined in black) remained at baseline fluorescence. The measured RFU values were consistent with those obtained when the same PCR assay was performed manually by a scientist, indicating that PRISM-generated protocols produce biologically valid results comparable to conventional hand-executed experiments.

% \ian{About the following. 
% 1) It's a nice example. 
% 2) Boiku et al. did something similar, we should cite them \cite{boiko2023autonomous}.
% 3) It seems lame to say ``not executed'' as that seems to imply that our automated lab is non-functional. 
% 4) I would argue strongly that we should run all four examples in Boiku Fig 4, to show that we at least match their capabilities. Ideally we would do something more complex. One idea that occurred to me is taking a picture, pixelizing it to 8x8 or whatever pixels, then mixing colors to draw the picture. Wouldn't that be cool?
% 5) As noted above, we don't appear to describe the ``Cell Painting assay protocol'' except to assert that it is ``complex.''
% }

To evaluate the generality of PRISM beyond PCR, we used the framework to generate a Cell Painting assay protocol, as described in Section~\ref{sec:Use Case 3}. PRISM successfully produced a detailed set of robotic instructions covering reagent handling, plate setup, and imaging preparation. These instructions were validated in silico and by a domain expert. 

Together, these results confirm that PRISM can deliver end-to-end executable protocols within an automated laboratory and can also generalize beyond single-step molecular assays to more complex, multi-step biological workflows.

%% file: sections/conclusions.tex
\section{Conclusions}

We have presented PRISM, a framework that addresses the challenge of translating scientific hypotheses into error-free robotic instructions by coupling LLM-based planning systems with simulation-based validation. This integration ensures that protocols are vetted for physical reliability, enabling the application of generative AI within laboratory settings in which reliability concerns would otherwise prevent its use. We summarize in the following the system's performance across our key benchmarks, synthesize the insights gained regarding agentic reasoning, and outline the remaining challenges for fully self-driving laboratories.

\subsection{System Performance and Contributions}

We have demonstrated in this work the successful application of LLMs to automated protocol generation for laboratory research, establishing a framework that bridges the substantial gap between high-level experimental intent and executable robotic instructions. PRISM has shown compatibility with existing automated laboratory hardware by leveraging the MADSci platform, indicating that the approach does not require specialized or custom-built robotic systems to function effectively. Through the integration of simulation-based refinement via NVIDIA Omniverse and Isaac Sim, we have validated an approach that allows protocols to be iteratively improved before physical execution, reducing the risk of equipment damage and experimental failures.

We also reported on the results of a quantitative performance comparison across multiple LLMs. This study revealed distinct capabilities and limitations of different reasoning models in the context of laboratory automation. The system's effectiveness has been explicitly tied to real-world validation through successful PCR execution in our automated laboratory, while its generalizability has been demonstrated through in-silico generation of a more complex Cell Painting protocol.

\subsection{Key Findings and Insights}

The protocol planning evaluation revealed fundamental differences in how language models approach the decomposition of experimental procedures into executable steps. Multi-agent workflows demonstrated superior performance by enforcing explicit separation of concerns: retrieval of procedural knowledge, spatial reasoning about deck layout and labware positioning, step-by-step formatting with physical constraints, and systematic validation of completeness and logical flow. This structured approach reduced false positives (unnecessary actions) and false negatives (missing critical steps) compared to single-agent reasoning, which tended to produce protocols that were either overly verbose or missing essential details. Notably, the iterative refinement mechanism—where the critique and validator agents provide feedback to the planner—proved essential for achieving convergence, with most models requiring 2 or 3 refinement cycles to reach acceptable accuracy levels. The comparison between constrained and open-ended prompting revealed that while constrained prompts improved initial generation quality by providing explicit guidance, open-ended prompts tested the models' ability to reason independently about experimental requirements, volumes, and transfer sequences. This finding suggests that protocol planning performance depends not only on model capability but also on the degree of structural scaffolding provided through prompt design and agent specialization.

Our evaluation revealed model-specific strengths for different aspects of protocol generation, with variations in both initial generation quality and error correction capabilities. The effectiveness of iterative refinement in achieving valid protocols was demonstrated across all tested models, though the improvement per-iteration varied significantly based on model reasoning capabilities. Weaker models required multiple iterations to correct individual instances of the same error, while stronger models could identify and fix all instances simultaneously. As shown in the ablation study, removal of simulation feedback reduced a medium-capability model's performance down to that of simulation-assisted low-capability models. Simulation proved essential for preventing real-world failures, enabling the reliable detection of physical execution issues such as collisions, missing opening steps for robots with specific operational constraints, and invalid command sequences that would have caused failures in the physical laboratory. Importantly, the PCR experiments showed comparable performance to manual execution, demonstrating that automation through LLM-driven protocol generation does not compromise biological outcomes and can achieve the same scientific results as traditional manual approaches.

\subsection{Current Limitations and Future Work}

Despite promising results, several limitations warrant attention in future work. LLMs still require substantial prompt engineering to achieve consistent and reliable performance, with sensitivity to prompt phrasing varying notably across model families. Developing adaptive prompting strategies or domain-specific fine-tuning approaches tailored to laboratory protocols could reduce this dependence on manual prompt design. %Additionally, c
Current models also continue to struggle with reasoning about multi-well operations, complex volume calculations, and many-to-many transfer patterns---limiting their ability to generalize to more sophisticated assays. Addressing these challenges will be critical for extending automated protocol generation to increasingly complex experimental workflows.
% Despite the promising results, several limitations warrant attention in future research. LLMs often require extensive prompt engineering to achieve reliable performance, with some models such as Mistral needing significantly more guidance compared to Claude. Future work should explore adaptive prompting strategies or fine-tuning approaches specifically tailored for laboratory contexts, which could reduce the engineering effort required to achieve consistent results. Current models struggle with reasoning about operations involving multiple wells, complex volume calculations, and many-to-many transfer patterns, which limits their ability to handle more sophisticated assays. Progress in this area will be essential for scaling the approach to increasingly complex experimental protocols.

The development of liquid world models represents another promising direction for future research. Such models could capture physical phenomena including viscosity effects, evaporation rates, and contamination risks that are currently absent from simulations, enabling more comprehensive validation of protocols before physical execution. Beyond text-based inputs, incorporating additional modalities such as images of experimental setups, CAD models of laboratory equipment, and real-time sensor data may improve the robustness and accuracy of protocol generation.

The current Omniverse simulation has scope limitations, in particular regarding physical motion validation through reachability and collision feedback. Simulation fidelity constraints also exist, such as the lack of liquid physics modeling and the absence of chemical or biological process simulation, which require that scientific accuracy must be verified through other means. Expanding Omniverse error detection capabilities to include scientific accuracy validation and more detailed protocol adherence verification would strengthen the system's ability to catch errors before physical execution.

We also note that current text-based simulation feedback limits intuitive understanding of complex protocol execution flows for errors that are not expressed clearly in textual output formats. Recent work such as Agentic Lab \cite{wang2025agentic} demonstrates the potential of vision-language models for laboratory reasoning, suggesting future directions for PRISM to incorporate multimodal simulation feedback—visualizing predicted equipment states, timing relationships, and potential failure modes alongside textual explanations to enhance protocol refinement. 

Finally, we observe that simulation environments such as that considered here could be leveraged for laboratory design optimization, with digital twin modeling used to help identify laboratory configurations that maximize efficiency and minimize potential collision or access conflicts among robots.

\subsection{Broader Impact}

The PRISM framework has the potential to democratize automated laboratory research by lowering the barriers to entry for institutions and researchers who lack specialized expertise in robotics and automation programming. By enabling scientists to describe experiments in structured natural language rather than requiring detailed knowledge of robot control systems, the PRISM approach makes laboratory automation accessible to a broader community. This democratization carries significant implications for reproducibility and standardization in biological protocols, as automatically generated protocols can be shared, validated, and executed consistently across different laboratories with compatible automation platforms.

The work presented here provides a pathway toward fully autonomous experimental design and execution, where the entire process from hypothesis formation through experimental execution and data analysis is automated. By bridging LLMs, physics-based simulation, and laboratory automation hardware, PRISM contributes to the foundation for self-driving laboratories. Such systems have the potential to accelerate scientific discovery while simultaneously reducing human error in protocol execution, enabling more reliable and reproducible experimental results. As these technologies mature and become more widely adopted, they may fundamentally transform how laboratory research is conducted, shifting the role of scientists from manual execution to higher-level experimental design and interpretation of results.

%% file: sections/appendix.tex
\onecolumn
\appendix
%\section*{Appendix}

\section{Protocol Planning (Stage 1)}\label{sec:Appendix Protocol Planning}
This appendix presents the prompts described in Section~\ref{sec:Protocol Planning} that PRISM uses to generate English-language plans, which are then converted into robot-compatible steps for the PCR protocol. It includes the prompts used for both open-ended and constrained generation, leveraging multi-agent and reasoning-model approaches.

\subsection{Multi-Agent Framework}
\vspace{0.6em}
\subsubsection{Constrained Prompting for PCR}
\rule{\linewidth}{1pt}
\vspace{0.3em}  % add space after the rule

\noindent\textbf{WebSurfer Agent Prompt}
{\small
\input{appendix/multi-agent/constrained/websurfer_prompt}
}
\vspace{0.8em}

% --- Protocol Planner ---
\noindent\textbf{Protocol Planner Prompt}
\vspace{0.3em}
{\small
\input{appendix/multi-agent/constrained/protocol_planner_prompt}
}
\vspace{0.8em}

% --- Critique Agent ---
\noindent\textbf{Critique Agent Prompt}
\vspace{0.3em}
{\small
\input{appendix/multi-agent/constrained/Critique_prompt}
}
\vspace{0.8em}

% --- Validator Agent ---
\noindent\textbf{Validator Agent Prompt}
\vspace{0.3em}
{\small
\input{appendix/multi-agent/constrained/Validator_prompt}
}
\vspace{1em}

\subsubsection{Open-Ended Prompting for PCR}
\rule{\linewidth}{1pt}
\vspace{0.3em}  % add space after the rule

\noindent\textbf{WebSurfer Agent Prompt}
{\small
\input{appendix/multi-agent/open-ended/PRISM_WebSurfer_Prompt}
}
\vspace{0.8em}

% --- Protocol Planner ---
\noindent\textbf{Protocol Planner Prompt}
\vspace{0.3em}
{\small
\input{appendix/multi-agent/open-ended/PRISM_Protocol_Planner_Prompt}
}
\vspace{0.8em}

% --- Critique Agent ---
\noindent\textbf{Critique and Validator Prompt}
\vspace{0.3em}
{\small
\input{appendix/multi-agent/open-ended/PRISM_Critique_Validator_Prompt_v2}
}
\vspace{1em}

\subsubsection{Cell Painting Prompts}
\rule{\linewidth}{1pt}
\vspace{0.3em}

% --- WebSurfer Agent ---
\noindent\textbf{WebSurfer Agent Prompt}\\[0.6em]
% \vspace{0.3em}
{\small
\input{appendix/cell-painitng/multi-agent/Websurfer_prompt_updated}
}
\vspace{1em}

% --- Protocol Planner Agent ---
\noindent\textbf{Protocol Planner Prompt}\\[0.6em]
% \vspace{0.3em}
{\small
\input{appendix/cell-painitng/multi-agent/Robotic_setup_prompt_cellpainting}
}
\vspace{1em}

% --- Critique Agent ---
\noindent\textbf{Critique Agent Prompt}\\[0.6em]
{\small
\input{appendix/cell-painitng/multi-agent/Critique_prompt_cellpainting}
}
\vspace{1em}

% --- Validator Agent ---
\noindent\textbf{Validator Agent Prompt}\\[0.6em]
{\small
\input{appendix/cell-painitng/multi-agent/Validator_prompt_cellpainting}
}
\vspace{1em}

\subsection{Single Agent}
\vspace{0.6em}
\subsubsection{Constrained Prompting}
\rule{\linewidth}{1pt}
\vspace{0.3em}  % add space after the rule
{\small
\input{appendix/reasoning/constrained/prompt_fixed}
}
\vspace{0.6em}
\subsubsection{Open-Ended Prompting}
\rule{\linewidth}{1pt}
\vspace{0.3em}  % add space after the rule
{\small
\input{appendix/reasoning/open-ended/open_ended}
}
\vspace{1em}
\subsubsection{Cell Painting Prompts}
\rule{\linewidth}{1pt}
\vspace{0.3em}

% --- WebSurfer Agent ---
% \noindent\textbf{WebSurfer Agent Prompt}\\[0.6em]
% \vspace{0.3em}
{\small
\input{appendix/cell-painitng/single-agent/prompt}
}
\vspace{1em}

% \twocolumn
\clearpage
\section{Protocol Generation (Stage 2)}
\subsection{Initial Protocol Generation Prompt}
The following is the full prompt used in Section~\ref{sec:Protocol Generation / Validation} for PRISM to convert the English-language plans into robot-compatible steps for the PCR protocol. It covers the available robots in the laboratory, the desired output formats, and ends with the specific workflow to be converted.

\noindent
\rule{\linewidth}{1pt}
%\vspace{0.3em}  % add space after the rule
\input{appendix/simulation/protocol_generation_prompt}

\clearpage
\subsection{Initial Protocol Generation Output}
The following is the raw output from Claude Sonnet 4.5 on the prompt in the previous section. Note that the raw output describes two files (a Python file and a YAML file), which were manually extracted and used in the downstream tasks.

\noindent
\rule{\linewidth}{1pt}
%\vspace{0.3em}  % add space after the rule
\input{appendix/simulation/protocol_generation_output}

\clearpage
\subsection{Iteration with Simulation Feedback Prompt}
The following is the full response given to the Claude Sonnet 4.5 model based on the simulation results of the previous section's output, which prompts Claude to identify and fix the error. This process is repeated with each new error until the simulation detects no errors.

\noindent
\rule{\linewidth}{1pt}
%\vspace{0.3em}  % add space after the rule
\input{appendix/simulation/iteration_prompt}

%% file: appendix/multi-agent/constrained/websurfer_prompt.tex
\begin{spverbatim}

You are the first agent (WebSurfer) in the PRISM workflow, responsible for gathering comprehensive details about the Luna Universal qPCR Master Mix Protocol.

Your objective is to get the Luna Universal qPCR Master Mix protocol and extract:
– reagent list and volumes  
– step-by-step liquid handling instructions  
– thermocycling conditions (3 step version)  
– final absorbance measurement using a Hidex reader.

Your output from this should be clear English instructions, structured as:
1. Materials
2. Liquid Handling Steps
3. Thermocycling Steps
4. Absorbance Reading

Avoid melt-curve analysis and keep all numbers explicit (no ranges).
\end{spverbatim}

%% file: appendix/multi-agent/constrained/protocol_planner_prompt.tex
\begin{spverbatim}

You are the second agent (Protocol Planner) in the PRISM workflow, following the WebSurfer agent.
Your role is to convert the constants gathered by the WebSurfer into a complete, executable OT-2 liquid-handling and thermocycling protocol for the Luna Universal qPCR Master Mix.

### Reagent Plate Layout (Reaction Plate)
A1–A2: Luna Universal qPCR Master Mix  
C1–C4: Nuclease-free Biowater  
D1: Forward Primer  
E1: Reverse Primer  
H1: Template DNA  

### Constants
– Reaction wells (A1, A3) include Template DNA.  
– Control wells (A2, A4) exclude Template DNA.  
– Volumes (µL): qPCR Mix 20, Biowater 5, Forward Primer 5, Reverse Primer 5, Template DNA 5.
***Notę: Biowater should be used to make up the 40 µL reaction volume in control wells  
- 3 thermocycling steps

### Output Format
List all steps in plain English using this exact format:
`X.) Transfer [volume] of [reagent] from [Well on plate] on reaction plate to [destination well] on final plate with 3 mix cycles. [Tip action - eject tip]`

After those steps, append:
`N.) Thermocycling: [Temperature] for [Duration]`

Ensure steps are logically ordered, distinct per well, and ready for direct OT-2 execution.
\end{spverbatim}

%% file: appendix/multi-agent/constrained/Critique_prompt.tex
\begin{spverbatim}
You are the third agent (Critique) in the PRISM workflow, following the Protocol Planner.
Your role is to review the Planner’s generated OT-2 protocol for the Luna Universal qPCR Master Mix and provide specific, actionable feedback to correct any physical, logical, or formatting errors.

Your objective is to analyze the PCR protocol generated by the Robotic agent and check for,

– Verify reagent volumes and wells match the following specifications:
- Volumes (µL): qPCR Mix 20, Biowater 5, Forward Primer 5, Reverse Primer 5, Template DNA 5.
*** Notę: Biowater should be used to make up the 40 µL reaction volume in control wells
– Reaction wells (A1, A3) include Template DNA; control wells (A2, A4) exclude it.  
– 3 thermocycling steps are present.  
– Each transfer step follows this exact format:
  `X.) Transfer [volume] of [reagent] from [Well on plate] on reaction plate to [destination well] on final plate with 3 mix cycles. [Tip action - eject tip]`
– All reagent source wells match the reagent plate layout:
  A1–A2 = Luna Master Mix, C1–C4 = Biowater, D1 = Forward Primer, E1 = Reverse Primer, H1 = Template DNA.
– Steps are logically ordered and feasible for the OT-2.

Provide concise, numbered feedback listing any formatting errors, incorrect volumes, wrong source wells, or unnecessary steps.

\end{spverbatim}

%% file: appendix/multi-agent/constrained/Validator_prompt.tex
\begin{spverbatim}

You are the fourth agent (Validator) in the PRISM workflow, following the Critique agent.
Your objective is to verify that the final qPCR protocol is accurate, fully corrected, and ready for OT-2 execution.

Your objective is to confirm that the final PCR protocol is accurate, formatted correctly, and executable on the OT-2 system.

validation criteria is,
– All reagent volumes and well assignments match the constants.  
– 3 thermocycling steps are present.  
– Every step follows the required format:
  `X.) Transfer [volume] of [reagent] from [Well on plate] on reaction plate to [destination well] on final plate with 3 mix cycles. [Tip action - eject tip]`
– Reagent source wells correspond exactly to:
  A1–A2 = Luna Master Mix, C1–C4 = Biowater, D1 = Forward Primer, E1 = Reverse Primer, H1 = Template DNA.  
– Logical order, minimal motion, and incorporation of all Critique feedback.  

Return only the final corrected and validated protocol, formatted for direct execution by the OT-2 robot.

\end{spverbatim}

%% file: appendix/multi-agent/open-ended/PRISM_WebSurfer_Prompt.tex
\begin{spverbatim}

You are the WebSurfer agent, part of the PRISM scientific workflow which uses a multi-agent framework to generate an OT-2–executable Luna® Universal qPCR protocol.

Objective: Retrieve the official Luna Universal qPCR Master Mix protocol from the vendor and extract only what’s needed to run a minimal qPCR on the OT‑2.

Return a single, self-contained report with these sections in plain English:

CONSTANTS (SOURCE-OF-TRUTH)
- Per-reaction component volumes (µL): Luna Master Mix, nuclease-free water, Forward Primer, Reverse Primer, Template DNA.
- Total reaction volume (µL).
- Acceptable reaction volume range and any notes about scaling.
- Fluorescent dye/readout and compatible detection notes (channel/filter if stated).
- Mixing/handling cautions relevant to automation (e.g., recommended mixing, bubble avoidance).
- Thermocycling profile expressed in exactly three steps (initial step; cycling step with temperatures, durations, and cycle count; final step). Use °C and s only; no ranges.

MATERIALS
- Reagents, labware, instruments.

LIQUID HANDLING GUIDANCE
- Vendor notes that influence robotic transfers, mixing, or order of addition.

THERMOCYCLING STEPS (3 step TEST VARIANT)
- Derive a 3 step version directly from the vendor profile (do not invent temperatures or times).

FLUORESCENCE READING
- Instrument/dye compatibility and readout channel. If the vendor does not specify, write exactly “Not specified by vendor.”

Rules:
- Make every numeric value explicit (µL, °C, s). No ranges.
- If the vendor omits a fact, say “Not specified by vendor.”
- Do not include melt-curve analysis unless explicitly stated by vendor.

\end{spverbatim}

%% file: appendix/multi-agent/open-ended/PRISM_Protocol_Planner_Prompt.tex
\begin{spverbatim}

You are the Protocol Planning agent, part of the PRISM scientific workflow which uses a multi-agent framework to generate an OT-2–executable Luna® Universal qPCR protocol. Your input is the WebSurfer report above.

GOAL
Produce a complete, physically feasible, biologically correct Luna qPCR plan for an Opentrons OT‑2 that will be consumed by downstream agents.

YOU MUST
1) Assign reagent and sample wells.
2) Design both the reagent plate (sources) and the final plate (PCR reactions) layouts.
3) Generate fully enumerated, strictly formatted OT‑2 liquid handling steps.
4) Include exactly 3 thermocycling steps (initial, cycling, final) using WebSurfer’s CONSTANTS.
5) Add a final fluorescence read on the Hidex.

DECK & LABWARE (declare explicitly)
- final plate (PCR reactions): slot 1
- reagent plate (sources): specify slots but be explicit
- tip racks: specify slots but be explicit
- pipettes: single‑channel P20 and/or P300; select the smallest suitable for the protocol
- use fresh tips for every transfer

VOLUME & ACCURACY RULES
- Treat WebSurfer’s CONSTANTS as the source‑of‑truth.
- Minimum accurate transfer = 5 µL. If any specified single transfer would be < 5 µL (e.g., 3 µL Forward Primer or 3 µL Reverse Primer) it will be inaccurate transfer. Try to keep the minimum possible volume 5ul.
- Scale all components proportionally while maintaining valid concentrations (e.g., preserve the master mix factor).
- If you change total reaction volume (e.g., to 40 µL), verify and state that concentrations remain compliant with CONSTANTS.
- Prepare each source on the reagent plate with >=10% overage and SHOW THE MATH.
- Keep the working volume of each well on the plate in mind while designing the reagent plate (e.g., 96-well PCR plate can hold 100ul volume in each well and if you require >= 100ul of any reagent you should have 2 wells on the plate with the reagent).

FINAL PLATE DESIGN
- Exactly 2 Test wells (include Template DNA) and 2 Control wells (exclude Template DNA). Identify explicitly (e.g., A1, A3 = Test; A2, A4 = Control).
- Keep the wells apart on final reaction plate to avoid any contamination.

STRICT STEP FORMAT (no shorthand, no “repeat for wells” language)
 X.) Transfer [volume] of [reagent] from [Well on plate] on reaction plate to [destination well] on final plate with 3 mix cycles. [Tip action - eject tip]

OUTPUT STRUCTURE (and nothing else)
1) Materials & Plate Map
2) Liquid Handling Steps (strictly formatted)
3) Thermocycling Steps (3 steps)
4) Fluorescence Read (Hidex; state channel if specified by WebSurfer; otherwise write “Channel not specified by vendor.”)

\end{spverbatim}

%% file: appendix/multi-agent/open-ended/PRISM_Critique_Validator_Prompt_v2.tex
\begin{spverbatim}

You are the combined Critique + Validator agent, part of the PRISM scientific workflow which uses a multi-agent framework to generate an OT-2–executable Luna Universal qPCR protocol.

Inputs:
- Protocol Planner’s output (Materials & Plate Map, Liquid Handling Steps, Thermocycling, Fluorescence Read)
- WebSurfer report (CONSTANTS — treat this as the source-of-truth)

--------------------------------
CRITIQUE — CHECKS (enumerate issues with concrete fixes)
--------------------------------

A) CONSISTENCY WITH CONSTANTS
1. Per-reaction component volumes match WebSurfer CONSTANTS OR are proportionally scaled with concentrations preserved (e.g., maintain 2× master mix factor).
2. Total per-well volume equals the declared total.
3. Any scaling is explicitly justified and keeps primer/template concentrations valid.

B) PLATE MAPS, OVERAGE & CAPACITY
4. Reagent plate lists all sources with >=10% overage and shows arithmetic.
5. Reagent well capacity constraints are respected: if any reagent requires >=100 µL total on a 96‑well PCR plate, it must be split across >=2 source wells with explicit mapping.
6. Final plate has exactly 2 Test wells (include Template DNA) and 2 Control wells (exclude Template DNA), and the chosen wells are non‑adjacent to reduce contamination risk. Identify them explicitly (e.g., A1, A3 = Test; A2, A4 = Control).

C) TRANSFER MINIMUM, ENUMERATION & TIP HYGIENE
7. No single transfer is < 5 µL. If WebSurfer constants specify sub‑5 µL components (e.g., 3 µL primers), confirm proportional scaling or other compliant solution that keeps concentrations valid.
8. Every liquid‑handling instruction is fully enumerated — no “repeat for wells” shorthand.
9. Fresh tips used when touching primers or template; pipette choice is appropriate for the protocol (e.g., P20 vs P300).

D) FORMAT, LABWARE & DECK
10. **Strict line-format compliance (must match EXACTLY):**
    X.) Transfer [volume] of [reagent] from [Well on plate] on reaction plate to [destination well] on final plate with 3 mix cycles. [Tip action - eject tip]
    - Volume unit must be µL; reagent names must match Materials/Plate Map.
    - Source wells must exist on the reagent plate; destination wells must exist on the final plate.
11. Deck declarations are explicit and consistent: final plate = slot 1; reagent plate = slot 2; tip racks = slots 7–8 (or explicitly noted alternatives).

E) THERMOCYCLING & HIDEX
12. Exactly 3 thermocycling steps, matching CONSTANTS in °C and s (no ranges).
13. Hidex read step present; channel specified if in CONSTANTS, else the text “Channel not specified by vendor.”

If all checks pass: return “No issues found.”
Otherwise: enumerate issues and provide corrected lines and/or corrected sections.

--------------------------------
VALIDATOR — FINAL OUTPUT
--------------------------------
After applying fixes, return ONLY the final corrected protocol with these sections:
1) Materials & Plate Map
2) Liquid Handling Steps
3) Thermocycling Steps
4) Fluorescence Read

STRICT LIQUID-HANDLING LINE FORMAT (must match EXACTLY)
X.) Transfer [volume] of [reagent] from [Well on plate] on reaction plate to [destination well] on final plate with 3 mix cycles. [Tip action - eject tip]
- Replace placeholders with concrete values (e.g., “40 µL”, “Luna Universal qPCR Master Mix”, “A1”, etc.).
- Do not use shorthand or “repeat steps” language.

\end{spverbatim}

%% file: appendix/cell-painitng/multi-agent/Websurfer_prompt_updated.tex
\begin{spverbatim}
You are a highly capable agent tasked with surfing the internet to gather comprehensive details about the PhenoVue™ Cell Painting Kit and its protocol. Your objective is to extract and clearly present:

A detailed list of materials included in the PhenoVue™ Cell Painting Kit.
Step-by-step instructions for preparing staining solutions using the kit reagents.
Protocol steps for cell preparation, staining, fixation, permeabilization, and washing.
Specific guidelines for reagent volumes, incubation times, and handling conditions (e.g., temperature, light sensitivity).
Recommendations for imaging parameters (e.g., microscope settings for fluorescence channels).
Ensure all information is presented in clear, straightforward English. Avoid technical jargon and highlight details essential for robotic automation.
\end{spverbatim}

%% file: appendix/cell-painitng/multi-agent/Robotic_setup_prompt_cellpainting.tex
\begin{spverbatim}
You are an advanced agent tasked with converting a biologically accurate Cell Painting protocol into a fully OT-2-compatible liquid handling protocol.

Your responsibilities:

1. Convert each step into OT-2-friendly format using:
   - Specific OT-2 slot numbers (1–11)
   - Valid labware types compatible with OT-2 (e.g., 12-column reservoirs, tube racks, 96-well plates)
   - Well positions within labware (e.g., A1, B2, C3)
   - Mix cycles (always 3)
   - Tip actions ("[Tip action: eject]")

2. Labware and labeling:
   - Assign each labware item (e.g., a 96-well plate or reservoir) a **clear, generic label** like “reaction plate,” “stain reservoir,” or “DNA tube rack.”
   - These labels must be unique and **referenced consistently** throughout the protocol steps.
   - If multiple items of the same labware type are used (e.g., two 96-well plates), assign them distinct labels and slot numbers.

3. Reagent and material tracking:
   - Before the protocol steps, generate a **List of Reagents** and a **List of Materials**, with their assigned OT-2 slots and well positions.
   - In the Materials section, list each labware item with:
     - Slot number
     - Labware type (e.g., 12-channel reservoir, 96-well plate)
     - Assigned label (e.g., “reaction plate”)

4. Reagent volume management:
   - If a reagent is used in multiple steps, reuse the same well until its volume is exhausted.
   - When needed, switch to the next available well or column for that reagent.
   - Track cumulative usage to ensure sufficient volume remains.

Format (MANDATORY) for all pipetting actions:

    X.) Transfer [volume] of [reagent] from [labware label] well [source well] to [labware label] well [destination well] with 3 mix cycles. [Tip action: eject tip]

Examples:

    1.) Transfer 100 µl of Fixation Buffer from stain reservoir well A3 to reaction plate well B1 with 3 mix cycles. [Tip action: eject]
    2.) Transfer 50 µl of 1X HBSS from wash reservoir well C2 to reaction plate well D2 with 3 mix cycles. [Tip action: eject]

Notes:
- Do mention labware type and label (e.g., “tube rack,” “reservoir,” “reaction plate”) in the step.
- Do not refer to slots directly in the steps — use the assigned label from the Materials list.
- All steps must be robot-ready, numbered, and fully self-contained.

Your final output must include:
1. A List of Reagents (reagent name + well position + labware label)
2. A List of Materials (slot + labware type + label)
3. A numbered list of OT-2 liquid handling steps (robot-ready and formatted)

This protocol must be biologically accurate, physically feasible, and fully executable on OT-2.
\end{spverbatim}

%% file: appendix/cell-painitng/multi-agent/Critique_prompt_cellpainting.tex
\begin{spverbatim}
You are tasked with critically analyzing the Cell Painting protocol prepared by the Robotic Agent. Your objective is to ensure the protocol's:

- Formatting compliance
- Logical correctness
- Efficiency and optimization for robotic execution
- Proper use of labware labels and reagent/material tracking

Step Formatting:

Each liquid handling instruction must follow this exact format:

    X.) Transfer [volume] of [reagent] from [labware label] well [source well] to [labware label] well [destination well] with 3 mix cycles. [Tip action: eject tip]

Example:
    1.) Transfer 50 µl of 1X HBSS from stain reservoir well A1 to reaction plate well B3 with 3 mix cycles. [Tip action: eject tip]

Invalid Examples:
    - "Add 100 µl of PBS" (missing structure and identifiers)
    - "Transfer cells to new plate" (non-specific, no volumes or positions)
    - "From plate A1 to plate B1" (no reagent name, no labware label)

Your critique tasks:

1. **Reagent and Material Lists**
   - Does the protocol include a complete **List of Reagents** and **List of Materials**?
   - Are reagents properly assigned to wells within labeled labware?
   - Are materials assigned to OT-2 slots with valid labware names and unique labels?

2. **Step Validation**
   - Are all steps properly numbered and formatted?
   - Are reagent names, well positions, and labware labels included and consistent?

3. **Protocol Logic and Flow**
   - Does the sequence follow biological logic for Cell Painting (e.g., staining before fixation)?
   - Are incubation, sealing, and unsealing instructions placed logically?

4. **Optimization Suggestions**
   - Is there unnecessary repetition or redundant transfers?
   - Could the protocol improve pipette efficiency or reduce tip usage?
   - Are reagents reused logically before switching to another well?

Important:
- The format must match robotic standards and be biologically sound.
- Flag issues clearly with bullet points or a numbered list.
- Use robotic-style examples when suggesting improvements.
\end{spverbatim}

%% file: appendix/cell-painitng/multi-agent/Validator_prompt_cellpainting.tex
\begin{spverbatim}
You are the final validation checkpoint for the Cell Painting protocol generated by the Robotic Agent.

You must confirm the protocol is:
- Robot-compatible
- Logically and structurally sound
- Using valid labware and formatting

---

REQUIRED STRUCTURE:

1. A **List of Reagents** with:
   - Reagent names
   - Assigned labware label
   - Well positions (e.g., A1, B2)

2. A **List of Materials** with:
   - OT-2 slot number
   - Labware type (e.g., 12-column reservoir, 96-well plate)
   - Assigned label (e.g., stain reservoir, reaction plate)

3. A list of **numbered pipetting steps**, each following this format:

    X.) Transfer [volume] of [reagent] from [labware label] well [source well] to [labware label] well [destination well] with 3 mix cycles. [Tip action: eject tip]

Examples:
    1.) Transfer 100 µl of Fixation Buffer from stain reservoir well A3 to reaction plate well B1 with 3 mix cycles. [Tip action: eject tip]

Invalid:
    - Steps missing labware labels
    - Vague instructions (e.g., "add buffer to plate")
    - Steps without reagent names or well positions
    - Steps referencing slot numbers instead of labware labels

Your validation tasks:

1. **Format Enforcement**
   - Every step must follow the correct format above.
   - No slot numbers should appear in steps — use labware labels only.

2. **Protocol Logic**
   - Step order should match Cell Painting protocol (reagent prep → staining → fixation → washing → imaging)
   - Confirm use of sealing, unsealing, and incubation instructions

3. **Consistency Checks**
   - Every reagent and material referenced in steps must appear in the reagent/material list.
   - Reagents must consistently use the same well until exhausted, then shift to next available well.

Output Instructions:
- If all steps and structure are valid, return the full protocol
- If any requirement is unmet, return:

	- Protocol rejected: Step formatting, logic, or structure is invalid. Please regenerate.

Do not return partial results or explanations. Only validate or reject.
\end{spverbatim}

%% file: appendix/reasoning/constrained/prompt_fixed.tex
\begin{spverbatim}
You are a highly capable reasoning model responsible for generating and verifying a complete, executable qPCR protocol for the Luna Universal qPCR Master Mix, including both reaction and control wells.

Your objective is to:
1. Gather accurate reagent information and thermocycling details.
2. Generate the step-by-step liquid handling instructions for an Opentrons OT-2 robot.
3. Critically review your own output for accuracy, formatting, and logical feasibility.
4. Return the final corrected and validated protocol ready for execution.

---

### Reagent Plate Layout (Reaction Plate)
A1–A2: Luna® Universal qPCR Master Mix  
C1–C4: Nuclease-free Biowater  
D1: Forward Primer  
E1: Reverse Primer  
H1: Template DNA  

---

### Required Specifications
- Reaction wells: A1 and A3 (include Template DNA)  
- Control wells: A2 and A4 (exclude Template DNA)  
- Volumes (µL):  
  • qPCR Master Mix – 20  
  • Biowater –  5 for test and 10 for control 
  • Forward Primer – 5  
  • Reverse Primer – 5  
  • Template DNA – 5 (only for A1, A3)  
- Total: **18 liquid handling steps** + **3 thermocycling steps**  
- Use a 20 µL pipette for all transfers.

---

### Step Format
Each liquid handling step must strictly follow:
`N.) Transfer [volume] of [reagent] from [Well on plate] on reaction plate to [destination well] on final plate with 3 mix cycles. [Tip action - eject tip]`

Thermocycling steps must follow:
`N.) Thermocycling: [Temperature] for [Duration].`

---

### Self-Check Before Final Output
After generating the protocol, verify:
- All reagent volumes and well mappings are correct.
- Reaction wells include Template DNA; control wells exclude it.
- Exactly 18 liquid-handling and 3 thermocycling steps exist.
- Every step follows the exact required format.
- Steps are ordered logically and feasible for OT-2 execution.

---

### Output Format
Produce only the final validated protocol, structured as:

1. Materials  
2. Liquid Handling Steps  
3. Thermocycling Steps  
4. Absorbance Reading (using Hidex plate reader)

Avoid melt-curve analysis. Keep all numbers explicit (no ranges). Do not include commentary or reasoning in the final answer.

\end{spverbatim}

%% file: appendix/reasoning/open-ended/open_ended.tex
\begin{spverbatim}
You are a highly capable reasoning model responsible for producing a complete, OT-2–executable qPCR protocol using the Luna Universal qPCR Master Mix. 
Your job is to (A) retrieve authoritative protocol constants, and (B) convert them into an OT-2–compatible protocol that fits the lab setup below. Produce only the final validated protocol. The generated protocol must be **physically feasible on an OT-2** and **biologically sane** (maintain correct concentrations, avoid inhibitory conditions, and minimize cross-contamination).

GOALS
1) Retrieve the official Luna Universal qPCR Master Mix protocol and extract required constants: component list, per-reaction volumes, thermocycling temperatures/times, and dye compatibility.
2) Generate explicit, enumerated OT-2 liquid-handling and thermocycling steps that satisfy the LAB SETUP & CONSTRAINTS.
3) Include two TEST and two CONTROL reactions placed far apart to minimize cross-contamination.
4) End with a fluorescence read on the Hidex plate reader.

LAB SETUP & CONSTRAINTS
- OT-2 deck: **final PCR plate** in **Slot 1**, **reagent plate** in **Slot 2**.  
- Use **96-well PCR plates** for both.  
- **Working volume per well:** 100 µL. If total reagent use exceeds this, split it across multiple source wells and note it in the Plate Map.  
- **Minimum single aspiration/dispense:** 5 µL — no transfers below this volume.  
- **Forward and reverse primers must be handled separately.** Each primer requires its own transfer step (>= 5 µL) per reaction well; 
- **Reaction volume for TEST wells:** 40 µL (derived from Luna constants).  
- **CONTROL wells:** identical setup but omit template DNA (replace with proper diluent).  
- **Thermocycling:** exactly 3 steps total using Luna’s official stage temperatures/durations.  
- **Fluorescence read:** Hidex plate reader, SYBR/FAM channel.  
- Do **not** use “repeat steps”; enumerate every transfer explicitly.  
- Assign source wells yourself and report them in the Plate Map.  
- Include any necessary overage/dead-volume adjustments in the reagent table.

FORMAT RULES
Produce only these six sections in order — no commentary:
  1. Materials  
  2. Plate Maps  
  3. Reagent Preparation (include overage math if used)  
  4. Liquid Handling Steps  
  5. Thermocycling Steps  
  6. Fluorescence Read  

Each liquid-handling instruction must follow this exact format:  
  N.) Transfer [volume] of [reagent] from [source well] on reagent plate to [destination well] on final plate with 3 mix cycles. [Tip action – eject tip]

Thermocycling format:  
  N.) Thermocycling: [TEMPERATURE °C] for [DURATION s].  

Fluorescence read (one line):  
  N.) Fluorescence read: Hidex plate reader, SYBR/FAM channel.

PLANNING REQUIREMENTS (INTERNAL – DO NOT PRINT)
- Compute per-component volumes from official Luna constants so TEST = 40 µL.
- Choose two TEST and two CONTROL wells with maximal spatial separation; list them in the Plate Map.  
- If any reagent’s total use exceeds 100 µL, split across multiple source wells and record mapping.  
- Verify 5 µL minimum per aspiration, proper reagent-to-well mappings, correct inclusion/exclusion of template, and formatting compliance.  
- All steps must be executable with standard OT-2 pipettes (fresh tip per transfer, correct mixing).  
- Output must contain exactly the six sections above and nothing else.

\end{spverbatim}

%% file: appendix/cell-painitng/single-agent/prompt.tex
\begin{spverbatim}
You are a single high-level reasoning model responsible for generating a complete, executable Cell Painting protocol using the PhenoVue™ Cell Painting Kit, designed for Opentrons OT-2 automation.

You will think and act holistically — not as multiple agents — to plan, reason through, and output the final validated protocol ready for robotic execution.

OBJECTIVE: 

Produce a self-contained OT-2-ready protocol that is biologically correct, physically feasible, and formatted for robotic execution.

TASKS

1. UNDERSTAND THE TASK
   - Recall or infer all necessary details about the PhenoVue™ Cell Painting Kit: reagents, staining and fixation solutions, incubation steps, and imaging readout.
   - Use reasoning to fill in realistic but clearly stated assumptions for reagent volumes, incubation times, or handling steps if exact data are not available.

2. DESIGN THE ROBOTIC WORKFLOW
   - Assign OT-2 labware to slots (1–11) and label them clearly (e.g., “reaction plate,” “stain reservoir,” “wash reservoir”).
   - Plan the sequence of operations following the logical Cell Painting workflow:
     cell seeding → staining → fixation → permeabilization → washing → imaging.
   - Ensure all transfers, incubations, and sealing/unsealing actions follow a practical robotic order.

3. GENERATE THE STRUCTURED OUTPUT
   The final protocol must include:
   - List of Reagents: reagent name + labware label + well position(s)
   - List of Materials: OT-2 slot + labware type + assigned label
   - Liquid Handling Steps: numbered and formatted exactly as:

     X.) Transfer [volume] of [reagent] from [Well on plate] on reaction plate to [destination well] on final plate with 3 mix cycles. [Tip action - eject tip]

4. REASON THROUGH VALIDATION
   - Check your own output for:
     • Correct structure (Reagents, Materials, Steps)
     • Consistency between lists and steps
     • Logical biological order and feasibility
     • Proper format (no slot numbers in steps, only labels)
   - If an inconsistency is found, fix it before final output.

5. DELIVER ONLY THE FINAL, VALIDATED PROTOCOL
   - Do not describe your reasoning process.
   - Output only the final, fully verified protocol, ready for OT-2 execution.

EXPECTED OUTPUT FORMAT
1. List of Reagents
2. List of Materials
3. Liquid Handling Steps (numbered)
4. Incubation / Sealing / Imaging Notes (if relevant)
\end{spverbatim}

%% file: appendix/simulation/protocol_generation_prompt.tex
{\small
\begin{spverbatim}
# YAML Workflow File Documentation

## Modules and Actions

### Module: `biopf400`
**Purpose**: Robotic transfer system (main plate handling robot)

#### Access:
This module does not have its own locations, and instead facilitates transfers between all other locations.

#### Restrictions:
- `transfer` requires that a plate is present at the source location
- `transfer` requires that the target location is available/empty
- `transfer` can only occur between locations with the same plate rotation (both narrow or both wide)
- If a transfer needs to change plate orientation (narrow to wide or wide to narrow), it must first go through the exchange station as an intermediate step

#### Actions:
- **`transfer`**: Moves plates between different locations/stations with configurable positioning
  - **Arguments**:
    - `source`: Source location identifier
    - `source_approach`: Approach path identifier for source pickup
    - `source_plate_rotation`: Plate orientation at source (`wide` or `narrow`)
    - `target`: Target location identifier
    - `target_approach`: Approach path identifier for target placement
    - `target_plate_rotation`: Plate orientation at target (`wide` or `narrow`)

### Module: `ot2bioalpha`
**Purpose**: OT-2 liquid handling robot

#### Access:
**Locations:**
- `ot2bioalpha_deck1_wide` - OT-2 robot deck position 1 (wide)

**Approach Paths:**
- `safe_path_ot2bioalpha` - Safe approach path for OT-2 robot

#### Restrictions:
- `run_protocol` can only be executed when plates and consumables are properly loaded on the deck

#### Actions:
- **`run_protocol`**: Executes a liquid handling protocol using a specified protocol file
  - **Arguments**:
    - `files`: Dictionary containing protocol files
    - `protocol`: Path to the OT-2 protocol file (e.g., `payload.ot2_protocol`)

### Module: `bio_sealer`
**Purpose**: Plate sealing station

#### Access:
**Locations:**
- `sealer_nest` - Plate sealing station nest (narrow)

**Approach Paths:**
- `safe_path_sealer` - Safe approach path for sealing station

#### Restrictions:
- `seal` can only be executed when a plate is present in the sealer nest

#### Actions:
- **`seal`**: Seals plates (typically with adhesive film or heat seal)
  - **Arguments**: None (empty args)

### Module: `bio_biometra3_96`
**Purpose**: Biometra thermocycler for PCR

#### Access:
**Locations:**
- `bio_biometra3_nest` - Biometra thermocycler nest (wide)

**Approach Paths:**
- `safe_path_biometra3` - Safe approach path for Biometra thermocycler

#### Restrictions:
- Must be `open` before a plate can be placed in or removed from the nest
- Must be `close`d before `run_program` can be executed
- `run_program` can only be executed when closed and with a plate present in the nest
- Must be `open`ed again after `run_program` completes to remove the plate

#### Actions:
- **`open`**: Opens the thermocycler lid
  - **Arguments**: None (empty args)
- **`close`**: Closes the thermocycler lid
  - **Arguments**: None (empty args)
- **`run_program`**: Executes a pre-programmed thermocycling protocol
  - **Arguments**:
    - `program_number`: Integer specifying which stored program to run (e.g., `5`)

### Module: `bio_peeler`
**Purpose**: Plate peeling/unsealing station

#### Access:
**Locations:**
- `peeler_nest` - Plate peeling station nest (narrow)

**Approach Paths:**
- `safe_path_peeler` - Safe approach path for peeling station

#### Restrictions:
- `peel` can only be executed when a sealed plate is present in the peeler nest

#### Actions:
- **`peel`**: Removes seals from plates
  - **Arguments**: None (empty args)

### Module: `hidex_geraldine`
**Purpose**: Hidex plate reader for assays/measurements

#### Access:
**Locations:**
- `hidex_geraldine_high_nest` - Hidex plate reader high nest position (narrow)

**Approach Paths:**
- `safe_path_hidex` - Safe approach path for Hidex plate reader

#### Restrictions:
- Must be `open` before a plate can be placed in or removed from the nest
- Must be `close`d before `run_assay` can be executed
- `run_assay` can only be executed when closed and with a plate present in the nest
- Must be `open`ed again after `run_assay` completes to remove the plate

#### Actions:
- **`open`**: Opens the plate reader lid
  - **Arguments**: None (empty args)
- **`close`**: Closes the plate reader lid
  - **Arguments**: None (empty args)
- **`run_assay`**: Runs a specified assay protocol
  - **Arguments**:
    - `assay_name`: String name of the assay protocol (e.g., `"PCR_Final_Results"`)

### Exchange Station
**Purpose**: The exchange station serves as an intermediate transfer point between modules, allowing for plate orientation changes and temporary storage during multi-step workflows.

#### Access:
**Locations:**
- `exchange_deck_high_wide` - Exchange station high position (wide)
- `exchange_deck_high_narrow` - Exchange station high position (narrow)

**Approach Paths:**
- `safe_path_exchange` - Safe approach path for exchange station

#### Restrictions:
- The exchange deck must either be a source or target location in a transfer, NEVER both
- The used exchange deck orientation must match the other location, being either both wide or both narrow
- Transfers via the exchange deck should use 2 transfers, the first having the exchange as the target, and the second having the exchange as the source

# YAML Workflow File Format Specification

This section describes the required format and structure for YAML workflow files used in automated laboratory protocols.

## Top-Level Structure

Every YAML workflow file must contain the following top-level fields followed by a `flowdef` section:

```yaml
name: [Protocol Name]
author: [Author/Organization]
info: [Description of the protocol]
version: '[Version Number]'

flowdef:
  # List of actions goes here
```

## Metadata Fields

### `name` (Required)
- **Type**: String
- **Purpose**: Human-readable name of the protocol
- **Example**: `"Test Protocol"`, `"PCR Amplification Workflow"`

### `author` (Required)
- **Type**: String
- **Purpose**: Name of the person or organization that created the protocol
- **Example**: `"Autoprotocol"`, `"Lab Automation Team"`

### `info` (Required)
- **Type**: String
- **Purpose**: Brief description of what the protocol accomplishes
- **Example**: `"A PCR protocol written by Autoprotocol"`, `"Automated qPCR workflow for gene expression analysis"`

### `version` (Required)
- **Type**: String (quoted)
- **Purpose**: Version identifier for protocol tracking and updates
- **Example**: `'0.1'`, `'1.2.3'`, `'2024.01'`

## Flow Definition (`flowdef`)

The `flowdef` section contains an ordered list of actions that define the workflow execution sequence.

### Action Structure

Each action in the `flowdef` list must have the following structure:

```yaml
  - action: [action_name]
    name: [human_readable_description]
    module: [module_identifier]
    args: [arguments_object]        # Optional
    files: [files_object]           # Optional
```

### Action Fields

#### `action` (Required)
- **Type**: String
- **Purpose**: Specifies which action to perform
- **Valid Values**: See module documentation for available actions
- **Examples**: `run_protocol`, `transfer`, `seal`, `open`, `close`, `run_program`, `peel`, `run_assay`

#### `name` (Required)
- **Type**: String
- **Purpose**: Human-readable description of what this specific action does
- **Example**: `"Run liquid protocol"`, `"Transfer the destination plate from OT-2 to exchange"`

#### `module` (Required)
- **Type**: String
- **Purpose**: Identifies which equipment module will execute the action
- **Valid Values**: `ot2bioalpha`, `biopf400`, `bio_sealer`, `bio_biometra3_96`, `bio_peeler`, `hidex_geraldine`

#### `args` (Optional)
- **Type**: Object/Dictionary
- **Purpose**: Contains arguments specific to the action being performed
- **Usage**: Required for actions that need parameters (e.g., transfer locations, program numbers)
- **Example**:
```yaml
    args:
      source: ot2bioalpha_deck1_wide
      target: exchange_deck_high_wide
      source_approach: safe_path_ot2bioalpha
      target_approach: safe_path_exchange
      source_plate_rotation: wide
      target_plate_rotation: wide
```

#### `files` (Optional)
- **Type**: Object/Dictionary
- **Purpose**: Specifies file references needed for the action
- **Usage**: Used primarily with `run_protocol` actions
- **Example**:
```yaml
    files:
      protocol: payload.ot2_protocol
```

## Complete Example

```yaml
name: Sample Protocol
author: PRISM
info: Automated workflow with sealing
version: '1.0'

flowdef:
  - action: run_protocol
    name: Execute liquid handling protocol
    module: ot2bioalpha
    files:
      protocol: payload.ot2_protocol

  - action: transfer
    name: Move plate from OT-2 to sealer
    module: biopf400
    args:
      source: ot2bioalpha_deck1_wide
      source_approach: safe_path_ot2bioalpha
      source_plate_rotation: wide
      target: sealer_nest
      target_approach: safe_path_sealer
      target_plate_rotation: narrow

  - action: seal
    name: Seal the reaction plate
    module: bio_sealer
    args: {}
```

## Important Notes

- **YAML Syntax**: Ensure proper indentation (2 spaces recommended) and valid YAML formatting
- **Action Order**: Actions execute sequentially in the order listed
- **Empty Args**: Use `args: {}` for actions that require the args field but take no parameters
- **Comments**: Use `#` for comments and section dividers to improve readability
- **String Quoting**: Quote version numbers and strings containing special characters

# OT-2 Liquid Handling File Format Specification

This document describes the required format and structure for Python protocol files used with the Opentrons OT-2 liquid handling robot.

## Top-Level Structure

Every OT-2 protocol file must contain the following components:

```python
requirements = {"robotType": "OT-2"}
from opentrons import protocol_api

metadata = {
    "protocolName": "PCR",                     # Human-readable protocol name
    "author": "PRISM",                         # Protocol author/creator
    "description": "PCR",                      # Brief protocol description
    "apiLevel": "2.12",                        # Opentrons API version
    "info": "A PCR protocol written by PRISM", # Additional information
    "name": "PCR",                             # Short protocol name
    "version": "1.0"                           # Version identifier
}

def run(protocol: protocol_api.ProtocolContext):
    # Protocol implementation
```

## Protocol Function

### Module Loading
```python
module = protocol.load_module("Temperature Module", "3")
deck["3"] = module.load_labware("nest_96_wellplate_100ul_pcr_full_skirt")
deck["3"].set_offset(x=1.0, y=1.4, z=5.7)
```

### Standard Labware Loading
```python
deck["1"] = protocol.load_labware("nest_96_wellplate_100ul_pcr_full_skirt", "1")
deck["1"].set_offset(x=0.9, y=-0.5, z=0.6)
```

### Tip Rack Loading
```python
deck["7"] = protocol.load_labware("opentrons_96_tiprack_20ul", "7")
deck["7"].set_offset(x=0.2, y=1.6, z=-0.8)
```

### Pipette Loading
```python
pipettes["left"] = protocol.load_instrument(
    "p20_single_gen2",
    "left",
    tip_racks=[deck["7"]]
)
```

### Standard Transfer Pattern

The basic pattern for a single transfer operation with mixing:

```python
pipettes["left"].pick_up_tip()
pipettes["left"].well_bottom_clearance.aspirate = 1
pipettes["left"].aspirate(20.0, deck["3"]["A1"])
pipettes["left"].well_bottom_clearance.dispense = 1
pipettes["left"].dispense(20.0, deck["1"]["B2"])
pipettes["left"].mix(3, 20, deck["1"]["B2"])
pipettes["left"].blow_out()
pipettes["left"].drop_tip()
```

### Operation Breakdown

1. **`pick_up_tip()`**: Retrieves a new tip from the tip rack
2. **`well_bottom_clearance.aspirate`**: Sets height above well bottom for aspiration (in mm)
3. **`aspirate(volume, location)`**: Draws liquid from specified well
4. **`well_bottom_clearance.dispense`**: Sets height above well bottom for dispensing (in mm)
5. **`dispense(volume, location)`**: Dispenses liquid into specified well
6. **`mix(repetitions, volume, location)`**: Mixes by aspirating and dispensing repeatedly
7. **`blow_out()`**: Expels any remaining liquid from the tip
8. **`drop_tip()`**: Discards the used tip

### Mix-Only Operation

For mixing without transfer:

```python
pipettes["left"].pick_up_tip()
pipettes["left"].well_bottom_clearance.aspirate = 1
pipettes["left"].well_bottom_clearance.dispense = 1
pipettes["left"].mix(10, 20, deck["1"]["B2"])
pipettes["left"].blow_out()
pipettes["left"].drop_tip()
```

## Minimal Working Example

```python
requirements = {"robotType": "OT-2"}
from opentrons import protocol_api

metadata = {
    "protocolName": "Simple Transfer",
    "author": "PRISM",
    "description": "Basic liquid transfer example",
    "apiLevel": "2.12",
    "info": "Demonstrates minimal OT-2 protocol structure",
    "name": "Simple Transfer",
    "version": "1.0"
}

def run(protocol: protocol_api.ProtocolContext):
    deck = {}
    pipettes = {}

    ################
    # load labware #
    ################
    deck["1"] = protocol.load_labware("nest_96_wellplate_100ul_pcr_full_skirt", "1")
    deck["2"] = protocol.load_labware("nest_96_wellplate_100ul_pcr_full_skirt", "2")
    deck["7"] = protocol.load_labware("opentrons_96_tiprack_20ul", "7")
    pipettes["left"] = protocol.load_instrument("p20_single_gen2", "left", tip_racks=[deck["7"]])

    ####################
    # execute commands #
    ####################
    # Transfer 10 µL from source to destination with mixing
    pipettes["left"].pick_up_tip()
    pipettes["left"].well_bottom_clearance.aspirate = 1
    pipettes["left"].aspirate(10.0, deck["2"]["A1"])
    pipettes["left"].well_bottom_clearance.dispense = 1
    pipettes["left"].dispense(10.0, deck["1"]["A1"])
    pipettes["left"].mix(3, 10, deck["1"]["A1"])
    pipettes["left"].blow_out()
    pipettes["left"].drop_tip()
```

## Important Notes

- **Tip Usage**: Each transfer operation uses a fresh tip (pick_up_tip at start, drop_tip at end)
- **Well References**: Use string notation for well positions (e.g., `"A1"`, `"B2"`)
- **Comments**: Use `#` for comments and section dividers to improve readability
- **Sequential Execution**: Commands execute in the order written

# PCR Workflow Sequence Overview

These are the high-level workflow steps for an automated PCR processing experiment.

The `ot2bioalpha` OT-2 is prepared with the 96-well plates `react_plate` on a Temperature Module at deck position 3 and `dest_plate` at deck position 1

1.) Transfer 20 uL of qPCR Master Mix from react_plate to dest_plate: A1→B2, A2→B11, A1→G2, A2→G11 with 3 mix cycles. [Tip action – eject]
2.) Transfer 5 uL of Nuclease-free Biowater from react_plate to dest_plate: C1→B2, C2→B11, C3→G2, C4→G11 with 3 mix cycles. [Tip action – eject]
3.) Transfer 5 uL of Forward Primer from react_plate D1 to dest_plate wells B2, B11, G2, and G11 with 3 mix cycles. [Tip action – eject]
4.) Transfer 5 uL of Reverse Primer from react_plate E1 to dest_plate wells B2, B11, G2, and G11 with 3 mix cycles. [Tip action – eject]
5.) Transfer 5 uL of Template DNA from react_plate H1 to dest_plate wells B2 and G2 with 3 mix cycles. [Tip action – eject]
6.) Transfer 5 uL of Nuclease-free Biowater from react_plate to dest_plate: C2→B11, C4→G11 with 3 mix cycles. [Tip action – eject]
7.) Mix 10X with pipette set to 20 uL in wells B2, G2, B11, and G11
8.) Seal the plate to prepare it for thermocycling
9.) Perform a thermocycling operation (program #5)
10.) Peel the plate to prepare it for reading
11.) Perform an absorbance reading using the Hidex plate reader (protocol "PCR_Final_Results")
12.) Open the Hidex plate reader lid after analysis completion

---

You are PRISM, an automated protocol designer for scientific workflows.
Using the provided reference material, create all OT-2 python files and workflow yaml files for the PCR workflow sequence.
Group all adjacent liquid transfer operations into a single execution of a liquid handling protocol on the OT-2 robot. Come up with a unique and descriptive name for each OT-2 protocol Python file.
All steps outside of liquid handling, including any in comments or described as manual, should be handled in the yaml file.
\end{spverbatim}
}

%% file: appendix/simulation/protocol_generation_output.tex
{\small
\begin{spverbatim}
Looking at the PCR workflow sequence, I'll create the necessary files. Steps 1-7 are liquid handling operations that should be grouped into a single OT-2 protocol, while steps 8-12 involve other modules in the YAML workflow.

# File 1: pcr_reaction_setup.ot2_protocol

```python
requirements = {"robotType": "OT-2"}
from opentrons import protocol_api

metadata = {
    "protocolName": "PCR Reaction Setup",
    "author": "PRISM",
    "description": "Complete PCR reaction assembly with master mix, primers, template DNA, and controls",
    "apiLevel": "2.12",
    "info": "Automated PCR reaction setup protocol",
    "name": "PCR Reaction Setup",
    "version": "1.0"
}

def run(protocol: protocol_api.ProtocolContext):
    deck = {}
    pipettes = {}

    ################
    # load labware #
    ################
    # Load Temperature Module with react_plate at position 3
    module = protocol.load_module("Temperature Module", "3")
    deck["3"] = module.load_labware("nest_96_wellplate_100ul_pcr_full_skirt")
    deck["3"].set_offset(x=1.0, y=1.4, z=5.7)

    # Load dest_plate at position 1
    deck["1"] = protocol.load_labware("nest_96_wellplate_100ul_pcr_full_skirt", "1")
    deck["1"].set_offset(x=0.9, y=-0.5, z=0.6)

    # Load tip rack
    deck["7"] = protocol.load_labware("opentrons_96_tiprack_20ul", "7")
    deck["7"].set_offset(x=0.2, y=1.6, z=-0.8)
    
    pipettes["left"] = protocol.load_instrument(
        "p20_single_gen2",
        "left",
        tip_racks=[deck["7"]]
    )

    ####################
    # execute commands #
    ####################
    
    # Step 1: Transfer 20 uL of qPCR Master Mix
    # A1→B2
    pipettes["left"].pick_up_tip()
    pipettes["left"].well_bottom_clearance.aspirate = 1
    pipettes["left"].aspirate(20.0, deck["3"]["A1"])
    pipettes["left"].well_bottom_clearance.dispense = 1
    pipettes["left"].dispense(20.0, deck["1"]["B2"])
    pipettes["left"].mix(3, 20, deck["1"]["B2"])
    pipettes["left"].blow_out()
    pipettes["left"].drop_tip()
    
    # A2→B11
    pipettes["left"].pick_up_tip()
    pipettes["left"].well_bottom_clearance.aspirate = 1
    pipettes["left"].aspirate(20.0, deck["3"]["A2"])
    pipettes["left"].well_bottom_clearance.dispense = 1
    pipettes["left"].dispense(20.0, deck["1"]["B11"])
    pipettes["left"].mix(3, 20, deck["1"]["B11"])
    pipettes["left"].blow_out()
    pipettes["left"].drop_tip()
    
    # A1→G2
    pipettes["left"].pick_up_tip()
    pipettes["left"].well_bottom_clearance.aspirate = 1
    pipettes["left"].aspirate(20.0, deck["3"]["A1"])
    pipettes["left"].well_bottom_clearance.dispense = 1
    pipettes["left"].dispense(20.0, deck["1"]["G2"])
    pipettes["left"].mix(3, 20, deck["1"]["G2"])
    pipettes["left"].blow_out()
    pipettes["left"].drop_tip()
    
    # A2→G11
    pipettes["left"].pick_up_tip()
    pipettes["left"].well_bottom_clearance.aspirate = 1
    pipettes["left"].aspirate(20.0, deck["3"]["A2"])
    pipettes["left"].well_bottom_clearance.dispense = 1
    pipettes["left"].dispense(20.0, deck["1"]["G11"])
    pipettes["left"].mix(3, 20, deck["1"]["G11"])
    pipettes["left"].blow_out()
    pipettes["left"].drop_tip()
    
    # Step 2: Transfer 5 uL of Nuclease-free Biowater
    # C1→B2
    pipettes["left"].pick_up_tip()
    pipettes["left"].well_bottom_clearance.aspirate = 1
    pipettes["left"].aspirate(5.0, deck["3"]["C1"])
    pipettes["left"].well_bottom_clearance.dispense = 1
    pipettes["left"].dispense(5.0, deck["1"]["B2"])
    pipettes["left"].mix(3, 20, deck["1"]["B2"])
    pipettes["left"].blow_out()
    pipettes["left"].drop_tip()
    
    # C2→B11
    pipettes["left"].pick_up_tip()
    pipettes["left"].well_bottom_clearance.aspirate = 1
    pipettes["left"].aspirate(5.0, deck["3"]["C2"])
    pipettes["left"].well_bottom_clearance.dispense = 1
    pipettes["left"].dispense(5.0, deck["1"]["B11"])
    pipettes["left"].mix(3, 20, deck["1"]["B11"])
    pipettes["left"].blow_out()
    pipettes["left"].drop_tip()
    
    # C3→G2
    pipettes["left"].pick_up_tip()
    pipettes["left"].well_bottom_clearance.aspirate = 1
    pipettes["left"].aspirate(5.0, deck["3"]["C3"])
    pipettes["left"].well_bottom_clearance.dispense = 1
    pipettes["left"].dispense(5.0, deck["1"]["G2"])
    pipettes["left"].mix(3, 20, deck["1"]["G2"])
    pipettes["left"].blow_out()
    pipettes["left"].drop_tip()
    
    # C4→G11
    pipettes["left"].pick_up_tip()
    pipettes["left"].well_bottom_clearance.aspirate = 1
    pipettes["left"].aspirate(5.0, deck["3"]["C4"])
    pipettes["left"].well_bottom_clearance.dispense = 1
    pipettes["left"].dispense(5.0, deck["1"]["G11"])
    pipettes["left"].mix(3, 20, deck["1"]["G11"])
    pipettes["left"].blow_out()
    pipettes["left"].drop_tip()
    
    # Step 3: Transfer 5 uL of Forward Primer from D1 to all destinations
    # D1→B2
    pipettes["left"].pick_up_tip()
    pipettes["left"].well_bottom_clearance.aspirate = 1
    pipettes["left"].aspirate(5.0, deck["3"]["D1"])
    pipettes["left"].well_bottom_clearance.dispense = 1
    pipettes["left"].dispense(5.0, deck["1"]["B2"])
    pipettes["left"].mix(3, 20, deck["1"]["B2"])
    pipettes["left"].blow_out()
    pipettes["left"].drop_tip()
    
    # D1→B11
    pipettes["left"].pick_up_tip()
    pipettes["left"].well_bottom_clearance.aspirate = 1
    pipettes["left"].aspirate(5.0, deck["3"]["D1"])
    pipettes["left"].well_bottom_clearance.dispense = 1
    pipettes["left"].dispense(5.0, deck["1"]["B11"])
    pipettes["left"].mix(3, 20, deck["1"]["B11"])
    pipettes["left"].blow_out()
    pipettes["left"].drop_tip()
    
    # D1→G2
    pipettes["left"].pick_up_tip()
    pipettes["left"].well_bottom_clearance.aspirate = 1
    pipettes["left"].aspirate(5.0, deck["3"]["D1"])
    pipettes["left"].well_bottom_clearance.dispense = 1
    pipettes["left"].dispense(5.0, deck["1"]["G2"])
    pipettes["left"].mix(3, 20, deck["1"]["G2"])
    pipettes["left"].blow_out()
    pipettes["left"].drop_tip()
    
    # D1→G11
    pipettes["left"].pick_up_tip()
    pipettes["left"].well_bottom_clearance.aspirate = 1
    pipettes["left"].aspirate(5.0, deck["3"]["D1"])
    pipettes["left"].well_bottom_clearance.dispense = 1
    pipettes["left"].dispense(5.0, deck["1"]["G11"])
    pipettes["left"].mix(3, 20, deck["1"]["G11"])
    pipettes["left"].blow_out()
    pipettes["left"].drop_tip()
    
    # Step 4: Transfer 5 uL of Reverse Primer from E1 to all destinations
    # E1→B2
    pipettes["left"].pick_up_tip()
    pipettes["left"].well_bottom_clearance.aspirate = 1
    pipettes["left"].aspirate(5.0, deck["3"]["E1"])
    pipettes["left"].well_bottom_clearance.dispense = 1
    pipettes["left"].dispense(5.0, deck["1"]["B2"])
    pipettes["left"].mix(3, 20, deck["1"]["B2"])
    pipettes["left"].blow_out()
    pipettes["left"].drop_tip()
    
    # E1→B11
    pipettes["left"].pick_up_tip()
    pipettes["left"].well_bottom_clearance.aspirate = 1
    pipettes["left"].aspirate(5.0, deck["3"]["E1"])
    pipettes["left"].well_bottom_clearance.dispense = 1
    pipettes["left"].dispense(5.0, deck["1"]["B11"])
    pipettes["left"].mix(3, 20, deck["1"]["B11"])
    pipettes["left"].blow_out()
    pipettes["left"].drop_tip()
    
    # E1→G2
    pipettes["left"].pick_up_tip()
    pipettes["left"].well_bottom_clearance.aspirate = 1
    pipettes["left"].aspirate(5.0, deck["3"]["E1"])
    pipettes["left"].well_bottom_clearance.dispense = 1
    pipettes["left"].dispense(5.0, deck["1"]["G2"])
    pipettes["left"].mix(3, 20, deck["1"]["G2"])
    pipettes["left"].blow_out()
    pipettes["left"].drop_tip()
    
    # E1→G11
    pipettes["left"].pick_up_tip()
    pipettes["left"].well_bottom_clearance.aspirate = 1
    pipettes["left"].aspirate(5.0, deck["3"]["E1"])
    pipettes["left"].well_bottom_clearance.dispense = 1
    pipettes["left"].dispense(5.0, deck["1"]["G11"])
    pipettes["left"].mix(3, 20, deck["1"]["G11"])
    pipettes["left"].blow_out()
    pipettes["left"].drop_tip()
    
    # Step 5: Transfer 5 uL of Template DNA from H1 to sample wells (not controls)
    # H1→B2
    pipettes["left"].pick_up_tip()
    pipettes["left"].well_bottom_clearance.aspirate = 1
    pipettes["left"].aspirate(5.0, deck["3"]["H1"])
    pipettes["left"].well_bottom_clearance.dispense = 1
    pipettes["left"].dispense(5.0, deck["1"]["B2"])
    pipettes["left"].mix(3, 20, deck["1"]["B2"])
    pipettes["left"].blow_out()
    pipettes["left"].drop_tip()
    
    # H1→G2
    pipettes["left"].pick_up_tip()
    pipettes["left"].well_bottom_clearance.aspirate = 1
    pipettes["left"].aspirate(5.0, deck["3"]["H1"])
    pipettes["left"].well_bottom_clearance.dispense = 1
    pipettes["left"].dispense(5.0, deck["1"]["G2"])
    pipettes["left"].mix(3, 20, deck["1"]["G2"])
    pipettes["left"].blow_out()
    pipettes["left"].drop_tip()
    
    # Step 6: Transfer 5 uL of Nuclease-free Biowater to control wells (instead of template)
    # C2→B11
    pipettes["left"].pick_up_tip()
    pipettes["left"].well_bottom_clearance.aspirate = 1
    pipettes["left"].aspirate(5.0, deck["3"]["C2"])
    pipettes["left"].well_bottom_clearance.dispense = 1
    pipettes["left"].dispense(5.0, deck["1"]["B11"])
    pipettes["left"].mix(3, 20, deck["1"]["B11"])
    pipettes["left"].blow_out()
    pipettes["left"].drop_tip()
    
    # C4→G11
    pipettes["left"].pick_up_tip()
    pipettes["left"].well_bottom_clearance.aspirate = 1
    pipettes["left"].aspirate(5.0, deck["3"]["C4"])
    pipettes["left"].well_bottom_clearance.dispense = 1
    pipettes["left"].dispense(5.0, deck["1"]["G11"])
    pipettes["left"].mix(3, 20, deck["1"]["G11"])
    pipettes["left"].blow_out()
    pipettes["left"].drop_tip()
    
    # Step 7: Final mixing - 10 cycles with 20 uL in all destination wells
    # Mix B2
    pipettes["left"].pick_up_tip()
    pipettes["left"].well_bottom_clearance.aspirate = 1
    pipettes["left"].well_bottom_clearance.dispense = 1
    pipettes["left"].mix(10, 20, deck["1"]["B2"])
    pipettes["left"].blow_out()
    pipettes["left"].drop_tip()
    
    # Mix G2
    pipettes["left"].pick_up_tip()
    pipettes["left"].well_bottom_clearance.aspirate = 1
    pipettes["left"].well_bottom_clearance.dispense = 1
    pipettes["left"].mix(10, 20, deck["1"]["G2"])
    pipettes["left"].blow_out()
    pipettes["left"].drop_tip()
    
    # Mix B11
    pipettes["left"].pick_up_tip()
    pipettes["left"].well_bottom_clearance.aspirate = 1
    pipettes["left"].well_bottom_clearance.dispense = 1
    pipettes["left"].mix(10, 20, deck["1"]["B11"])
    pipettes["left"].blow_out()
    pipettes["left"].drop_tip()
    
    # Mix G11
    pipettes["left"].pick_up_tip()
    pipettes["left"].well_bottom_clearance.aspirate = 1
    pipettes["left"].well_bottom_clearance.dispense = 1
    pipettes["left"].mix(10, 20, deck["1"]["G11"])
    pipettes["left"].blow_out()
    pipettes["left"].drop_tip()
```

# File 2: pcr_workflow.yaml

```yaml
name: Automated PCR Workflow
author: PRISM
info: Complete automated PCR workflow including reaction setup, thermocycling, and plate reading
version: '1.0'

flowdef:
  #########################
  # Liquid Handling Setup #
  #########################
  - action: run_protocol
    name: Execute PCR reaction setup protocol
    module: ot2bioalpha
    files:
      protocol: payload.pcr_reaction_setup.ot2_protocol

  ################################
  # Transfer to Sealer (via Exchange)
  ################################
  - action: transfer
    name: Transfer plate from OT-2 to exchange station
    module: biopf400
    args:
      source: ot2bioalpha_deck1_wide
      source_approach: safe_path_ot2bioalpha
      source_plate_rotation: wide
      target: exchange_deck_high_wide
      target_approach: safe_path_exchange
      target_plate_rotation: wide

  - action: transfer
    name: Transfer plate from exchange to sealer with orientation change
    module: biopf400
    args:
      source: exchange_deck_high_narrow
      source_approach: safe_path_exchange
      source_plate_rotation: narrow
      target: sealer_nest
      target_approach: safe_path_sealer
      target_plate_rotation: narrow

  ##########
  # Sealing
  ##########
  - action: seal
    name: Seal the PCR reaction plate
    module: bio_sealer
    args: {}

  ###################################
  # Transfer to Thermocycler (via Exchange)
  ###################################
  - action: transfer
    name: Transfer sealed plate from sealer to exchange station
    module: biopf400
    args:
      source: sealer_nest
      source_approach: safe_path_sealer
      source_plate_rotation: narrow
      target: exchange_deck_high_narrow
      target_approach: safe_path_exchange
      target_plate_rotation: narrow

  - action: transfer
    name: Transfer plate from exchange to thermocycler with orientation change
    module: biopf400
    args:
      source: exchange_deck_high_wide
      source_approach: safe_path_exchange
      source_plate_rotation: wide
      target: bio_biometra3_nest
      target_approach: safe_path_biometra3
      target_plate_rotation: wide

  ###############
  # Thermocycling
  ###############
  - action: open
    name: Open thermocycler lid
    module: bio_biometra3_96
    args: {}

  - action: close
    name: Close thermocycler lid
    module: bio_biometra3_96
    args: {}

  - action: run_program
    name: Execute PCR thermocycling program
    module: bio_biometra3_96
    args:
      program_number: 5

  - action: open
    name: Open thermocycler lid after completion
    module: bio_biometra3_96
    args: {}

  ###############################
  # Transfer to Peeler (via Exchange)
  ###############################
  - action: transfer
    name: Transfer plate from thermocycler to exchange station
    module: biopf400
    args:
      source: bio_biometra3_nest
      source_approach: safe_path_biometra3
      source_plate_rotation: wide
      target: exchange_deck_high_wide
      target_approach: safe_path_exchange
      target_plate_rotation: wide

  - action: transfer
    name: Transfer plate from exchange to peeler with orientation change
    module: biopf400
    args:
      source: exchange_deck_high_narrow
      source_approach: safe_path_exchange
      source_plate_rotation: narrow
      target: peeler_nest
      target_approach: safe_path_peeler
      target_plate_rotation: narrow

  ##########
  # Peeling
  ##########
  - action: peel
    name: Remove seal from PCR plate
    module: bio_peeler
    args: {}

  ####################################
  # Transfer to Plate Reader (via Exchange)
  ####################################
  - action: transfer
    name: Transfer peeled plate from peeler to exchange station
    module: biopf400
    args:
      source: peeler_nest
      source_approach: safe_path_peeler
      source_plate_rotation: narrow
      target: exchange_deck_high_narrow
      target_approach: safe_path_exchange
      target_plate_rotation: narrow

  - action: transfer
    name: Transfer plate from exchange to Hidex plate reader
    module: biopf400
    args:
      source: exchange_deck_high_narrow
      source_approach: safe_path_exchange
      source_plate_rotation: narrow
      target: hidex_geraldine_high_nest
      target_approach: safe_path_hidex
      target_plate_rotation: narrow

  ###############
  # Plate Reading
  ###############
  - action: open
    name: Open Hidex plate reader lid
    module: hidex_geraldine
    args: {}

  - action: close
    name: Close Hidex plate reader lid
    module: hidex_geraldine
    args: {}

  - action: run_assay
    name: Run PCR final results assay
    module: hidex_geraldine
    args:
      assay_name: "PCR_Final_Results"

  - action: open
    name: Open Hidex plate reader lid after analysis
    module: hidex_geraldine
    args: {}
```

These files implement the complete PCR workflow:

**pcr_reaction_setup.ot2_protocol**: Groups all liquid handling steps (1-7) into a single OT-2 protocol that:
- Adds qPCR Master Mix to all destination wells
- Adds nuclease-free water
- Adds forward and reverse primers to all wells
- Adds template DNA to sample wells (B2, G2)
- Adds water to control wells (B11, G11) instead of template
- Performs final mixing of all wells

**pcr_workflow.yaml**: Orchestrates the entire workflow including:
- Running the liquid handling protocol
- Transferring plates between modules using the exchange station for orientation changes
- Sealing, thermocycling, peeling, and plate reading operations
- Following all module restrictions (opening/closing lids, proper plate orientations)
\end{spverbatim}
}

%% file: appendix/simulation/iteration_prompt.tex
{\small
\begin{spverbatim}
Error while executing step 6 (`action=transfer, module=biopf400, args={'source': 'exchange_deck_high_wide', 'source_approach': 'safe_path_exchange', 'source_plate_rotation': 'wide', 'target': 'bio_biometra3_nest', 'target_approach': 'safe_path_biometra3', 'target_plate_rotation': 'wide'}`): Collision between microplate and bio_biometra3_96
Rewrite the full YAML file with the correction applied.
\end{spverbatim}
}